\documentclass[twosode, 11pt]{article}
\usepackage{caption}
\usepackage{subcaption}
\usepackage{arxiv_jmlr}
\usepackage[margin=1in]{geometry}
\usepackage{hyperref}
\usepackage{float}
\usepackage{color}
\usepackage{amssymb}
\usepackage{amsfonts}
\usepackage{amsxtra}     
\usepackage{srcltx}
\usepackage{graphicx}
\usepackage{verbatim}
\usepackage{algorithmicx}
\usepackage{algpseudocode}
\usepackage[]{algorithm2e}
\definecolor{blue}{rgb}{0,0,0.7}

\RestyleAlgo{boxruled}

\newcommand{\I}{\mbox{I}}

\newcommand{\X}{\mathbf{X}}
\newcommand{\x}{\mathbf{x}}
\newcommand{\parens}[1]{\left(#1\right)}
\newcommand{\cpr}{\mathbb{P}\left(y = 1 | \x\right)} 
\newcommand{\pr}{\mathbb{P}\left(\x \right)} 

\usepackage{epsfig}
\usepackage{enumerate}
\usepackage{color}

\begin{document}
\ShortHeadings{AdaBoost and Random Forests: the Power of Interpolation}{Wyner, Bleich, Olson, and Mease}
\firstpageno{1}
\title{Explaining the Success of AdaBoost and Random Forests as Interpolating Classifiers}

\author{\name Abraham J. Wyner \email  ajw@wharton.upenn.edu\\
		\name Matthew Olson \email maolson@wharton.upenn.edu \\
        \name Justin Bleich \email jbleich@wharton.upenn.edu \\
       \addr Department of Statistics\\
       Wharton School, University of Pennsylvania\\
       Philadelphia, PA 19104, USA
       \AND
       \name David Mease \email dmease@apple.com \\
        \addr Apple Inc.\\
	}
\editor{}

\maketitle

\begin{abstract}
There is a large literature explaining why  AdaBoost is a successful classifier.  The literature on AdaBoost focuses on classifier margins and boosting's interpretation as the optimization of an exponential likelihood function.  These existing explanations, however, have been pointed out to be incomplete. A random forest is another popular ensemble method for which there is substantially less explanation in the literature. We introduce a novel perspective on AdaBoost and random forests that proposes that the  two algorithms work for similar reasons.  While both classifiers achieve similar predictive accuracy, random forests cannot be conceived as a direct optimization procedure.  Rather, random forests is a self-averaging, interpolating algorithm which creates what we denote as a ``spiked-smooth" classifier, and we view AdaBoost in the same light.  We conjecture that both AdaBoost and random forests succeed because of this mechanism. We provide a number of examples to support this explanation. In the process, we question the conventional wisdom that suggests that boosting algorithms for classification require regularization or early stopping and should be limited to low complexity classes of learners, such as decision stumps.  We conclude that boosting should be used like random forests: with large decision trees,  without regularization or early stopping.
\end{abstract}

 
\section{ Introduction }
In the ``boosting'' approach to machine learning, a powerful ensemble of classifiers is formed by successively refitting a weak classifier  to different weighted realizations of a data set.  This intuitive procedure has seen a tremendous amount of success. In fact,  shortly, after its introduction, in a 1996 NIPS conference,  Leo Brieman crowned AdaBoost \citep{Freund1996}  (the first boosting algorithm)  the ``best off-the-shelf classifier in the world \citep{Friedman2000}."  AdaBoost's early success was immediately followed by efforts to explain and recast it in more conventional statistical terms. The statistical view of boosting holds that AdaBoost is a stage-wise optimization of an exponential loss function \citep{Friedman2000}. This realization was especially fruitful leading to new ``boosting machines" \citep{Friedman2001, Ridgeway2006} that could perform probability estimation and regression as well as adapt to different loss functions. The statistical view, however, is not the only explanation for the success of AdaBoost. The computer science literature has found generalization error guarantees using VC bounds from  PAC learning theory and margins \citep{guestrin2006}.  While some research has cast  doubt on the ability of any one of these to fully account for the performance of AdaBoost they are generally understood to be satisfactory \citep{Schapire2013}. 

	This paper parts with traditional perspectives on AdaBoost by concentrating our analysis on the  implications of the algorithm's ability to perfectly fit the training data in a wide variety of situations.  Indeed, common lore in statistical learning suggests that  perfectly fitting the training data must inevitably lead to ``overfitting."  This aversion is built into the DNA of a statistician who has been trained to believe, axiomatically, that  data can always be decomposed into signal and noise.  Traditionally,  the ``signal" is always modeled smoothly. The resulting residuals represent  the ``noise" or the random component in the data.  The statistician's art is to walk the balance between the signal and the noise, extracting as much signal as possible without extending the fit to  the noise.   In this light, it is counterintuitive that any classifier can ever be successful if every training example is  ``interpolated" by the algorithm and thus fit without error.
	
The computer scientist, on the other hand, does not automatically decompose problems into signal and noise.  In many classical problems, like image detection, there is no noise in the classical sense. Instead there are only complex signals. There are still residuals, but they do not represent irreducible random errors. If the task is to classify images into those with cats and without, the problem is hard not because it is noisy. There are no cats wearing dog disguises. Consequently, the computer scientist has no dogmatic aversion to interpolating training data.  This was the breakthrough.  

It is now well-known that interpolating classifiers can work, and work well.  The AdaBoost classifier created a huge splash by being better than its established competitors  (for instance, CART, neural networks, logistic regression) \citep{Breiman1998}  and substantively better than the technique of creating an ensemble using the bootstrap \citep{bagging}.  The statistics community was especially  confounded by two properties of AdaBoost: 1)  interpolation (perfect prediction in sample) was achieved after relatively few iterations, 2) generalization error continues to drop even after interpolation is achieved and maintained.

The main point of this paper is to demonstrate that AdaBoost and similar algorithms work not in spite,  but because of interpolation. To bolster this claim, we will draw a constant analogy with random forests \citep{Breiman2001}, another interpolating classifier. The random forests algorithm, which is also an ensemble-of-trees method, is generally regarded to be among the very best commonly used classifiers \citep{fernandez2014}.  Unlike AdaBoost, for which there are multiple accepted explanations, random forest's performance is much more mysterious since traditional statistical frameworks do not necessarily apply. The statistical view of boosting, for example, cannot apply  to random forests since the algorithm creates decision trees at random and then averages the results---there is no stage-wise optimization. In this paper, we will put forth the argument that both algorithms are effective for the same reason. We consider AdaBoost and random forests as canonical examples of ``interpolating classifiers," which we define to be a classifier's algorithmic property of fitting the training data completely without error.  Each of these interpolating classifiers also exhibits a self-averaging property.   We attempt to show that these two properties together make for a classifier with low generalization error. While it is easy to see that random forests has both of these mechanisms by design, it is less clear that this is true for AdaBoost.

It is worth noting that Breiman noticed the connection between random forests and AdaBoost as well, although his notion of a random forest was more general, including other types of large ensembles of randomly grown trees \citep{Breiman2001}.  In his 2001 \textit{Random Forests} paper, he conjectured that the weights of AdaBoost might behave like an ergodic dynamic system, converging to an invariant distribution.  When run for a long time, the additional rounds of AdaBoost were equivalent to drawing trees randomly grown according to this distribution, much like a random forest.  Recent work has followed up on this idea, proving that the weights assigned by AdaBoost do indeed converge to a invariant distribution \footnote{ More specifically, they consider the so called ``Optimal AdaBoost" algorithm, which is assume to pick the base classifier with lowest weighted error at each round.  They show that the per round average of any measurable function of the training weights converges under mild conditions.} \citep{belanich2012}.  In this work, the authors also show that functions of these weights, such as the generalization error and margins, also converge.  This work certainly complements ours, but we focus on the similarity between AdaBoost and random forests through the lens of the type of decision surfaces both classifiers produce, and ability of both algorithms to achieve zero error on the training set.
 
	One of our key contributions will be to present a decomposition of AdaBoost as the weighted sum of interpolating classifiers.  Another contribution will be to demonstrate the mechanism by which interpolation combined with averaging creates an effective classifier.  It turns out that interpolation provides a kind of robustness to noise: if a classifier fits the data extremely locally, a ``noise" point in one region will not affect the fit of the classifier at a nearby location.  When coupled with averaging, the result is that the fit stabilizes at regions of the data where there is signal, while the influence of noise points on the fit becomes even more localized. It will be easy to see this point holds true for random forests. For AdaBoost, it is less clear, however,  and a decomposition of AdaBoost and simulation results in Section~\ref{sec:avg_boost} will demonstrate this crucial point.  We will observe that the error of AdaBoost at test points near noise points will continue to decrease as AdaBoost is run for more iterations, demonstrating the localizing effect of averaging interpolating classifiers.
	
	We will begin in Section~\ref{sec:other_boost} by critiquing some of the existing explanations of AdaBoost.  In particular, we will discuss at length some of the shortcomings of the statistical optimization view of AdaBoost.  In Section~\ref{sec:interpolate}, we will discuss the merits of classification procedures that interpolate the training data, that is, that fit the training data set with no error.  The main conclusion from this section is that interpolation, done correctly, can provide robustness to a fit in the presence of noise.  This discussion will be augmented with simulations discussing the performance of random forests, AdaBoost, and other algorithms in a noisy environment.  We will then derive our central observation in Section~\ref{sec:avg_boost}, namely that AdaBoost can be decomposed as a sum of classifiers, each of which fits the training data perfectly.  The implication from this observation is that for the best performance, we should run AdaBoost for many iterations with deep trees.  The deep trees will allow the component classifiers to interpolate the data, while a large number of iterations will lend to a bagging effect.  We will then demonstrate this intuition in a real data example in Section ~\ref{sec:realData}. Finally, we conclude with a brief discussion in Section~\ref{sec:conclusion}.
	
\section{Competing Explanations for the Effectiveness of  Boosting}\label{sec:other_boost}

In this section we will present an overview of some of the most popular explanations for the success of boosting, with analysis of both the strengths and weaknesses of each approach.  Our emphasis will focus on the margins view of boosting and the statistical view of boosting, each of which has a large literature and has led to the development of variants of boosting algorithms.  For a more extensive review of the boosting literature, one is well-advised to consult \cite{schapire2012}.                          

    Before we begin, we will briefly review the AdaBoost algorithm not only to refresh the reader's mind, but also to establish the exact learning algorithm this paper will consider, as there are many variants of AdaBoost.  To this end, the reader is invited to review Algorithm ~\ref{algo:ada}.  In our setting, we are given $N$ training points $(\x_i,y_i)$ where $\x_i \in \mathcal{X}$ and $y_i \in \{-1,+1\}$.  On round $m$, where $m = 1, \ldots, M$, we fit a weak classifier $G_m\parens{\x}$ to a version of the data set reweighted by some weighting vector $\mathbf{w}_m$.  We then calculate the weighted misclassification rate of our chosen learner, and update the weighting measure used in the next round, $\mathbf{w}_{m+1}$.  The final classifier is output as the sign of a weighted linear combination of classifiers produced from each stage of the algorithm.  In practice, one sometimes limits the number of rounds of boosting as a form of regularization.  We will discuss this point more in the next section, and challenge its usefulness in later parts of the paper.
    
%
\begin{algorithm}[htp]
\caption{AdaBoost \citet{Hastie2009}}
\begin{algorithmic}
\State 1. Initialize the observation weights $w_i = \dfrac{1}{N},~~  i = 1, 2, \ldots, N$.

\State 2. For {$m=1$ to $M$}: 
\State \indent (a) Fit a classifier $G_m(\x)$ to the training data using weights $w_i$. \vspace{10pt}
\State\indent  (b) Compute $\mathrm{err}_m  = \dfrac{\sum_{i = 1}^N w_i I\parens{y_i \neq G_t\parens{\x_i}}}{\sum_{i = 1}^N w_i}$. \vspace{10pt}
\State\indent  (c) Compute $a_m = \log\parens{\dfrac{1- \mathrm{err}_t}{\mathrm{err}_t}}$.
\State\indent  (d) Set $w_i \leftarrow w_i \cdot \exp\parens{a_t \cdot I\parens{y_i \neq G_t\parens{\x_i}}}$ \vspace{15pt}
\State\indent  (e) Set $f_i(\x) = \sum_{m=1}^M a_m G_m\parens{\x}$ \vspace{15pt}
\State 3. Output $f(\x) = \mathrm{sign}\parens{f_M(\x)}$ 

\end{algorithmic}
\label{algo:ada}
\end{algorithm}

\subsection{Margin View of Boosting}
Some of the earliest attempts to understand AdaBoost's performance predicted that its generalization error would increase with the number of iterations: as AdaBoost is run for more rounds, it is able to fit the data increasingly well which should lead to overfitting. However, in practice we observe that running boosting for many rounds does not overfit in most cases.  One of the first  attempts to resolve this paradox was explored by \cite{Schapire1998}, who focused on the \textit{margins} of AdaBoost.  The margins can be thought of as a measure of how confident a classifier is about how it labels each point, and one would hypothetically desire to produce a classifier with as large of margins as possible.  \citet{Schapire1998} proved that AdaBoost's generalization error decreases as the size of the margins increase.  Indeed, in practice one observes that as AdaBoost is run for many iterations, test error decreases while the size of the empirical margins increase.   In fact, recent research has demonstrated that AdaBoost can be reformulated exactly as mirror descent applied to the problem of maximizing the smallest margin in the training set under suitable separability conditions \citep{Freund2013}.

One could take these observations to suggest that a more effective algorithm might be designed to explicitly optimize margins.  However, one can find evidence against this hypothesis in Breiman's \textit{arc-gv} algorithm \citep{breiman1999prediction}.  Breiman designed the \textit{arc-gv} algorithm to maximize the minimum margin in a data set, and he found that this algorithm actually had worse generalization error than AdaBoost.  Moreover, he developed generalization error bounds based on the minimum margin of a classifier which were tighter than those established by Shapire, casting doubt on the existing margin explanation.  Other algorithms designed to maximize margins, such as \textit{LP-Boost} have also been found to perform worse than AdaBoost in practice \citep{Wang2011}.  Critics of these supposed counterexamples to the margin view of boosting note that AdaBoost's success likely depends on the entire distribution of margins on the data, not just the smallest margin.  More recent work has improved upon the Breiman's generalization bound by taking into account other aspects of the margin that more closely reflect its distribution, adding new life to the margin explanation of AdaBoost \citep{gao2013}.  While the margin explanation of AdaBoost is certainly intuitive, its role in producing low generalization error is still an area of active research.

\subsection{Statistical Optimization View of Boosting}

\citet{Friedman2000} take great strides to clear up the mystery of boosting to provide statisticians with a statistical view of the subject. The heart of their article is the recasting of boosting as a statistically familiar program for finding an additive model by means of a forward stage-wise approximate optimization of an exponential criterion.  In short, this view places boosting firmly in classical statistical territory by clearly defining it as a procedure to search through the space of convex combinations of {\em weak} learners or {\em base} classifiers. This explanation has been widely assimilated and has reappeared in the statistical literature as well as in a plethora of computer science articles.    Subsequent to the seminal publication of \citet{Friedman2000} there has been a flurry of activity dedicated to theoretical analysis of the algorithm. This was made possible by the identification of boosting as optimization, which therefore admits of a mathematically tractable representation.  Research on the optimization properties of AdaBoost and the exponential loss function is still an active area of research, see \cite{Mukherjee2013}, for example.

Although the statistical optimization perspective of AdaBoost is surely interesting and informative, there remain problems.  First, we observe that the fact that AdaBoost minimizes an exponential loss may not alone account for its performance as a classifier.  \citet{Wyner2003} introduces a variant of AdaBoost called Beta-Boost which is very similar to AdaBoost except that by design the exponential loss function is constant throughout the iterations.  Despite this, Beta-Boost was able to demonstrate similar performance to AdaBoost on simulated data sets.  Furthermore, among many similar examples in the literature, \cite{Mease2008} present a simulation example in which the the exponential loss is monotonically increasing with the number of iterations of AdaBoost on a test set, while the generalization error decreases.  In this example, the value of the exponential loss is uninformative about how well the classifier generalizes.  \cite{Freund2013} also provide evidence to this end.  They conduct an experiment that compares AdaBoost to two AdaBoost variants that minimize the exponential loss function at differing rates: one performs the minimization very quickly through gradient descent, while the other performs the minimization quite slowly.  They find that AdaBoost performed significantly better than these two competitors, suggesting that AdaBoost's strong performance cannot be tied exclusively to its action on the exponential loss function.

We also contend that some of the mathematical theory connected with the statistical optimization view of boosting has a disconnect with the types of boosting algorithms that work in practice.  The optimization theory of boosting insists that overfitting can be avoided by requiring the set of weak learners, to be just that: \textit{weak}. \citet{Buhlmann2003} argues that one can avoid overfitting by employing regularization with weak base learners.  However, empirical evidence points to quite the opposite: boosting deep trees for many iterations tends to produce a better classifier than boosted stumps with regularization \citep{Mease2008}.  The use of early-stopping as a form of regularization has also been called into question \citep{Mease2008}.  The thrust of our paper will be to demonstrate why we \textit{should} actually expect boosting with deep trees run for many iterations to have better generalization error.  Recent work also suggests that boosting low complexity classifiers may not be able to achieve good accuracy in difficult classification tasks such as speech recognition or image recognition \citep{Cortes2014}.  This paper proposes an algorithm called ``DeepBoost" which encourages boosting high complexity base classifiers---such as very deep decision trees---but in a ``capacity-conscious" way.  One last problem with theory associated with the statistical view of boosting is that by its very nature it suggests that we should be able to extract conditional class probability estimates from the boosted fit, as the procedure is apparently maximizing a likelihood function.  \cite{Mease2008}, however, point out a number of examples where the implied conditional class estimates from the boosting fit diverge to zero and one.  While boosting appears to do an excellent job as a classifier, it apparently fails to estimate probability quantiles correctly.  

    We can now summarize the main empirical contradictions with existing theoretical explanations of boosting, which motivates the view we present in this paper:

\begin{enumerate}
\item Boosting works well, perhaps best in terms of performance if not efficiency, with ``strong learners" like C4.5 and CART \citep{niculescu2005}.
\item The value of exponential loss does not always bear a clear relationship to generalization error \citep{Mease2008}.
\item The optimization theory offers no explanation as to why the training error can be zero, yet the test error continues to descend \citep{freund2000}
\end{enumerate}

This paper will squarely depart from the statistical optimization view by asserting that AdaBoost may be best thought of as a (self) smoothed, interpolating classifier.  We will see that unlike the statistical optimization view, this perspective suggests that for best performance once should run many iterations of AdaBoost with deep trees.  This will allow us to draw a number of analogies between AdaBoost and random forests.  A key component to this argument will consist of explaining the success of interpolating classifiers in noisy environments.  We will pursue this line of thought in the following section.

 \section{Interpolating Classifiers}\label{sec:interpolate}

\begin{algorithm}[htp]
\caption{Random Forests \citet{Hastie2009}}
\begin{algorithmic}

\State 1. For {$b=1$ to $B$}: 
\State \indent (a) Draw a bootstrap sample $\X^*$ of size $N$ from the training data
\State \indent (b) Grow a decision tree $T_b$ to the data $\X^*$ by doing the following recursively \\ 
\indent \indent \indent until the minimum node size $n_{min}$ is reached:
\State \indent \indent i. Select $m$ of the $p$ variables
\State \indent \indent ii. Pick the best variable/split-point from the $m$ variables and partition \vspace{1pt}
\State 2. Output the ensemble $\{T_b\}_b^B$\\
\State  Let $\hat{C}_b(\x^*)$ be predicted class of tree $T_b$. Then $\hat{C}^B_{rf}(\x^*)=\mathrm{majority~vote}\{\hat{C}_b(\x^*)\}_1^B$.

\end{algorithmic}
\label{algo:rf}
\end{algorithm}

It is a widely held belief by statisticians that if a classifier interpolates all the data, that is, it fits all the training data without error,  then it cannot be consistent and should have a poor generalization error rate. In this section, we demonstrate that there are interpolating classifiers that defy this intuition: in particular, AdaBoost and random forests will serve as leading examples of such classifiers. We argue that these classifiers achieve good out of sample performance by maintaining a careful balance between the complexity required to perfectly match the training data and a general semi-smoothness property.  We begin with a quick review of the random forests classifier, which will be in constant analogy with AdaBoost.
 
\subsection{Random Forests}

Random forests has gained tremendous popularity due to robust performance across a wide range of data sets. The algorithm is often capable of achieving best-in-class performance with respect to low generalization error and is not highly sensitive to choice of tuning parameters, making it the off-the-shelf tool of choice for many applications. 

Algorithm~\ref{algo:rf} reviews the procedure for constructing a random forests model. Note that in many popular implementations, such as \texttt{R} implementation \texttt{randomForest} \citet{Liaw2002} built from Breiman's CART software, $n_{min}$ is set to one for classification.  This implies that each decision tree is designed to be grown to maximal depth and therefore necessarily interpolates the data in its bootstrap sample (assuming as least one continuous predictor). This results in each tree being a low bias but high variance estimator.  Variance is then reduced by averaging across trees, resulting in a ``smoothing'' of the estimated response surface. The random predictor selection within each tree further reduces variance by lowering the correlation across the trees.  The final random forest classifier still fits  the entire training data set  perfectly, at least with very high probability. To see this is true, consider any given training point. As the number of trees increases, with probability close to one,  that point will be present in the majority of the  bootstrap samples used to fit  the trees in the forest.  Thus the point will get the correct training set label when the votes are tabulated to determine the final class label. 

We wish to emphasize that despite its success, random forests is not directly optimizing any loss function across the entire ensemble;  each tree is grown independently of the other trees. While each tree may optimize a criteria such as the Gini index, the full ensemble is not constructed in any optimization-driven fashion such as is the case for AdaBoost. While there has been recent theoretical work describing the predictive surface of random forests \citep{wager2015}, the analysis required unnatural assumptions that are hard to justify in practice (such as the growth rate of minimum leaf size).  Rather, we postulate that the success of the algorithm is due to its interpolating nature plus the self-averaging mechanism.   We next consider the implications of interpolating classifiers more broadly.

\subsection{Local Robustness of Interpolating Classifiers}

Let us begin with a definition of interpolation:\\
\newline
{\bf Definition: }  Let $X_i$ be vector  observations of predictor variables a  let $Y_i$  be the observed  class label. A classifier $f(X)$  is said to be an {\em interpolating} classifier if for every training set example, the classifier assigns the correct class label; that is for every $i$, $f(X_i)=Y_i$. \\
\newline

The term ``interpolation" is likely jarring for some readers.  In many contexts, one often thinks about interpolating a set of points with classically smooth functions, such as polynomial splines.  However, strictly speaking, there are many other ways that one might interpolate a set of  points---through the fit of an AdaBoost classifier, for instance!  Since the notion of fitting a set of points without error is central to this paper, and since the common definition of interpolation does not preclude the kinds of fits we consider, we felt it appropriate to proceed with the term.

Many statisticians are not comfortable with classifiers that interpolate the training data: common wisdom suggests that any classifier which fits the training data perfectly must have poor generalization error.  Indeed, one of the first interpolating classifiers that might come to one's mind, the one-nearest neighbor, can be shown to be inconsistent and have poor generalization error in environments with noise.  Specifically, \cite{cover} have shown that the asymptotic generalization error for the one-nearest neighbor classifier is at least as large as the Bayes error rate.  However, the claim that all interpolating classifiers overfit is problematic, especially in light of the demonstrated success of classifiers that perfectly fit the training data, such as random forests.  

One of our key insights reverses  the common intuition about classifiers: interpolation can prevent overfitting.  An interpolated classifier, if sufficiently local, minimizes the influence of noise points in other parts of the data.  In order to make this point conceptually clear, it is helpful to put ourselves in familiar territory with a regression example.

\begin{figure}[htp]
\centering
\includegraphics[width=3.5in]{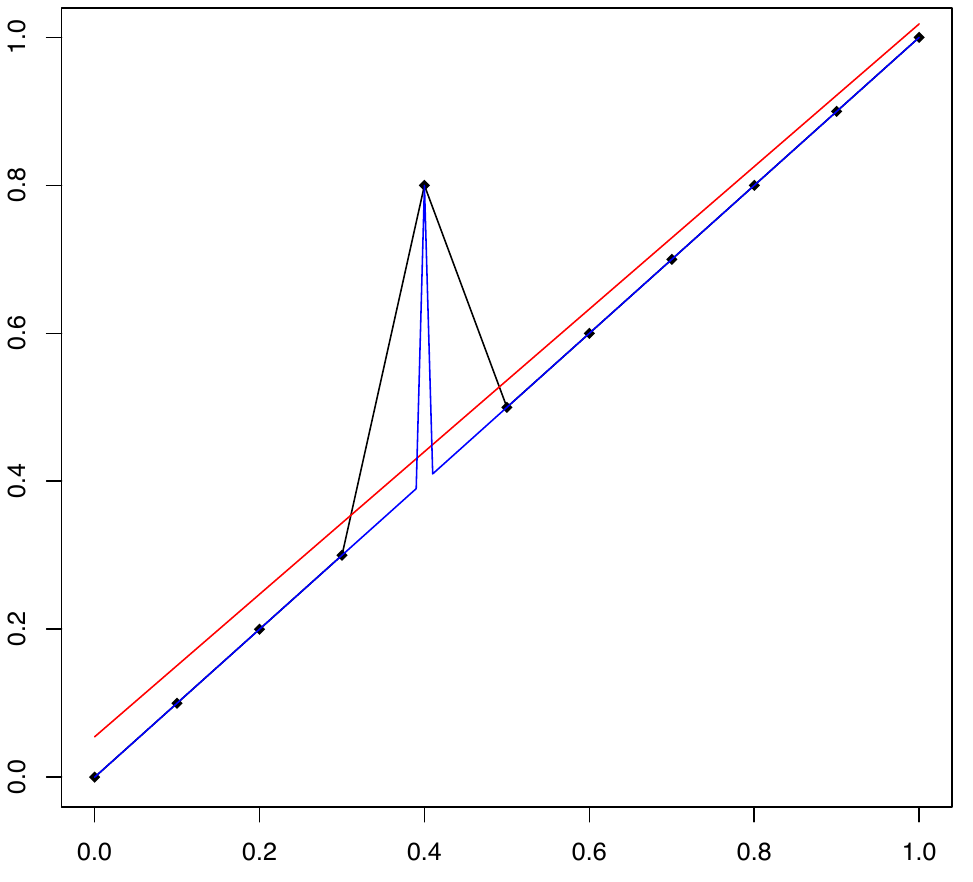}
\caption{Three estimated regression functions with two interpolating fits (black and blue) and an ordinary least squares fit (red).} 
\label{fig:smooth}
\end{figure}

Suppose we are trying to predict a continuous response $y$ based on real-valued $x$ observations. Let us assume that the true underlying model is $y = x + \epsilon$, where $\epsilon$ is a mixture of a point mass at zero and some heavy-tailed distribution.  In other words, we'll assume that most points in a given training set reflect the true linear relationship between $y $ and $x$, but a few observations will be noise points.  This is analogous to the types of probability models we typically consider in classification settings, such as those found in later sections of the paper.  Figure \ref{fig:smooth} shows hypothetical training data: note that the only ``noise" point is found at $x=0.4$.  We then consider fitting three models to this data: two interpolating functions, given by the blue and black lines, and an ordinary regression fit given by the red line.  The first thing to notice is that the two interpolating fits differ only from the true target mean model $y=x$ only at the noise point $x=0.4$.  In contrast, the fit of the regression line deviates from the underlying target over the entire range of $x$.  The one noise point corrupted the entire fit of the regression line, while the interpolating lines were able to minimize the influence of the noise point by adapting to it only very locally.  Moreover, one should note that between the two interpolating fits, the blue line interpolates more locally than the black line, and thus its overall fit is even less influenced by the noise point.  This simplified example is of course meant to be didactic, but we will show throughout the rest of this paper that in practice AdaBoost and random forest do indeed produce fits similar to the blue line.

\begin{figure}[htp]
\centering
    \begin{subfigure}[b]{0.48\textwidth}
        \includegraphics[width=\textwidth]{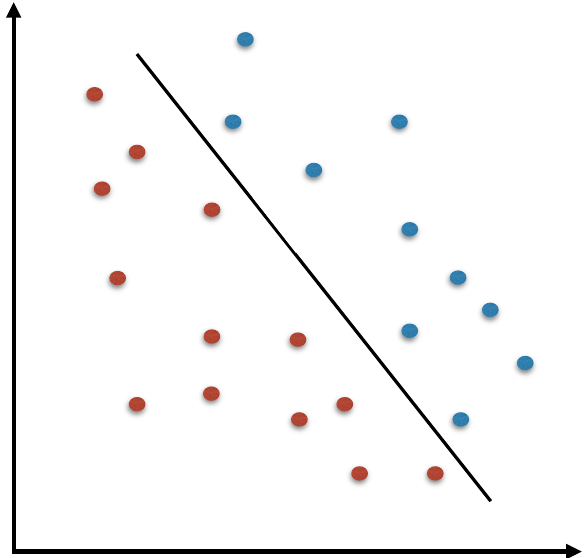}
        \caption{}
	\label{subfig:interp_a}
     \end{subfigure}
     \quad
     \begin{subfigure}[b]{0.48\textwidth}
        \includegraphics[width=\textwidth]{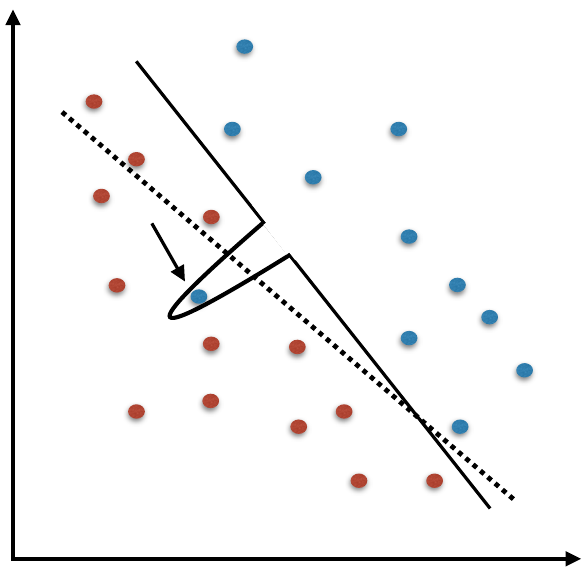}
        \caption{}
	\label{subfig:interp_b}
     \end{subfigure}
     \caption{An illustration of the robustness to noise of interpolating classifiers.   In \ref{subfig:interp_a} a classifier (such as a logistic regression) is fit to a set of training data, and the decision boundary produced by this classifier is shown as a solid line.  In \ref{subfig:interp_b}, the same type of classifier is fit to the training data, except a noise point is added to the data (the blue point marked by the arrow).  The one noise point shifts the entire decision boundary, which is shown as the dotted line.  On the other hand, the decision boundary produced by a classifier which interpolates very locally is shown as a solid line.  It is clear that this classifier is able to adapt locally to the noise point, and the overall fit does not get corrupted.}
          \label{fig:boundary_interp}
\end{figure}

While it is conceptually clear that it is desirable to produce fits like the blue interpolating line in the previous example, one may wonder how such fits can be achieved in practice.  In the classification setting, we will argue throughout this paper that this type of fit can be realized through the process of averaging interpolating classifiers.  We will refer to the decision surface produced by this process as being \textit{spiked-smooth}.  The decision surface is \textit{spiked} in the sense that it is allowed to adapt in very small regions to noise points, and it is \textit{smooth} in the sense that it has been produced through averaging.  A technical definition of a spiked-smooth decision surface would lead us too far astray.  It may be helpful instead to consider the types of classifiers that do not produce a spiked-smooth decision surface, such as a logistic regression.  Logistic regression separates the input space into two regions with a hyperplane, making a constant prediction on each region.  This surface is not spiked-smooth because it does not allow for local variations: in large regions (namely, half-spaces), the classifier's predictions are constrained to be the same sign.  Figure~\ref{fig:boundary_interp} provides a graphical illustration of this intuition.

Intuitively, we expect ensembles of interpolating classifiers to generalize well because they are flexible enough to fit a complex signal, and ``local" enough to prevent the undue influence of noise points. It is clear that random forests are averaged collections of interpolating classifiers, and we will show in section \ref{sec:avg_boost} that AdaBoost may be thought of in the same way. Classically smooth classifiers---such as logistic regression or pruned CART trees---are forced to produce fits that are locally constant.  It is harder for such classifiers to ``recover" from making a mistake at a noise point since the surface of the fit will affect the fit at nearby points.  Interpolating classifiers are flexible enough to make mistakes in ``small regions."  When averaged over many such classifiers, the influence of the noise point can be easily smoothed out.  Later examples in the paper will visualize this process, and will demonstrate that the later iterations of AdaBoost have a smoothing effect which shrinks down the influence of noise points on the overall fit of the classifier.

With this discussion in mind, let us consider another conceptual example, this time in the classification setting.  Suppose we have have two
predictors $x_1$ and $x_2$ distributed independently and uniformly
on $[0,1]^2$ and $y \in \{-1, +1\}$. Further suppose that the true conditional class
probability function is
\begin{equation*}
\pr=\cpr = p=.75
\end{equation*}
for all $\x$.  This is a pure noise model with no signal, but in
general one could view this as a subspace of a more complex model
in which the $\pr$ function is approximately constant.  Since the
Bayes decision rule is to classify every point as a ``+1'', we
would desire an algorithm that will match the Bayes rule as 
close as possible.  Again, we stress that this closeness
should be judged with respect to the population or a hold-out
sample.  On this training data,  any interpolating classifier will
necessarily  differ from the Bayes rule for $1-p=25\%$ of
the points on average.

\begin{figure}[htp]
 \centerline{\includegraphics[width=3.5in]{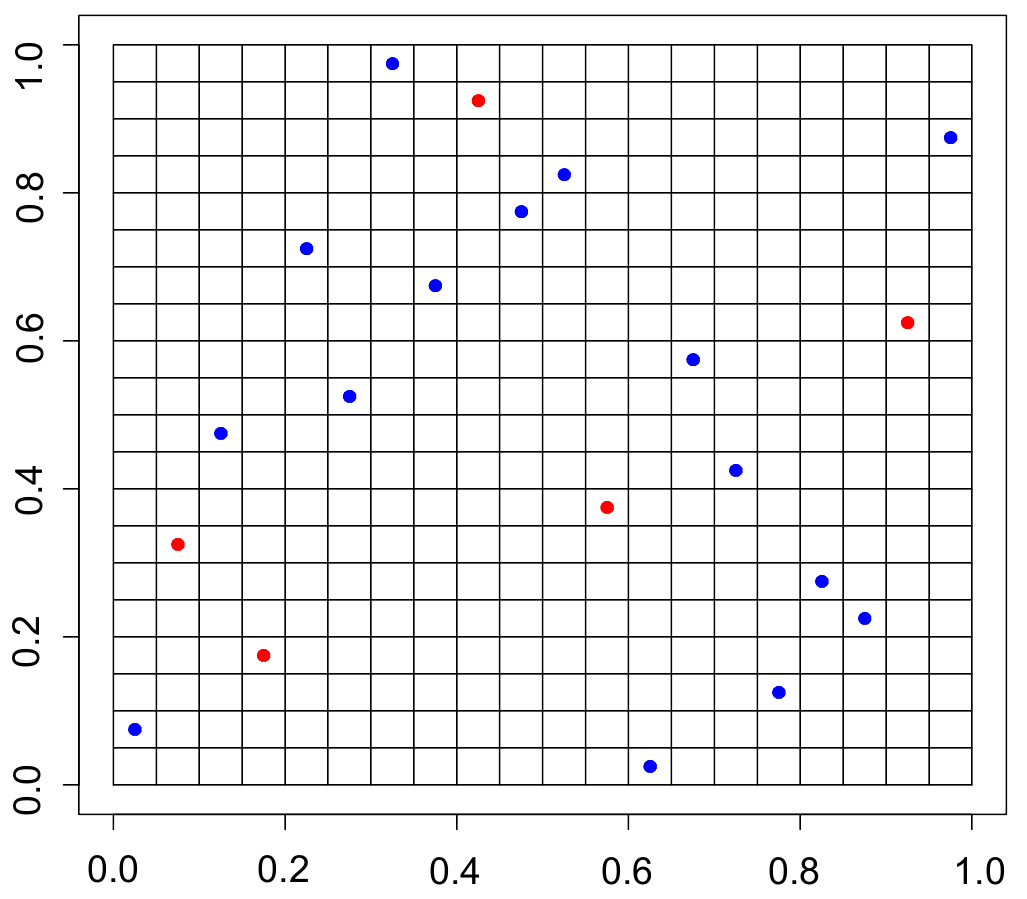}} \caption{Training Data} \label{fig:trainingdata}
\end{figure}

Figure \ref{fig:trainingdata} shows a possible sample of training data
from this model of size $n=20$. The blue points represent the
``+1'''s and the red points represent the ``-1"'s. There are
$5/20=25\%$ red points.  The training data was sampled according
to a Latin Hypercube design using the midpoints of the squares so
that the points would be evenly spaced, but that is not essential.

\begin{figure}[htp]
\centering
    \begin{subfigure}[b]{0.48\textwidth}
        \includegraphics[width=\textwidth]{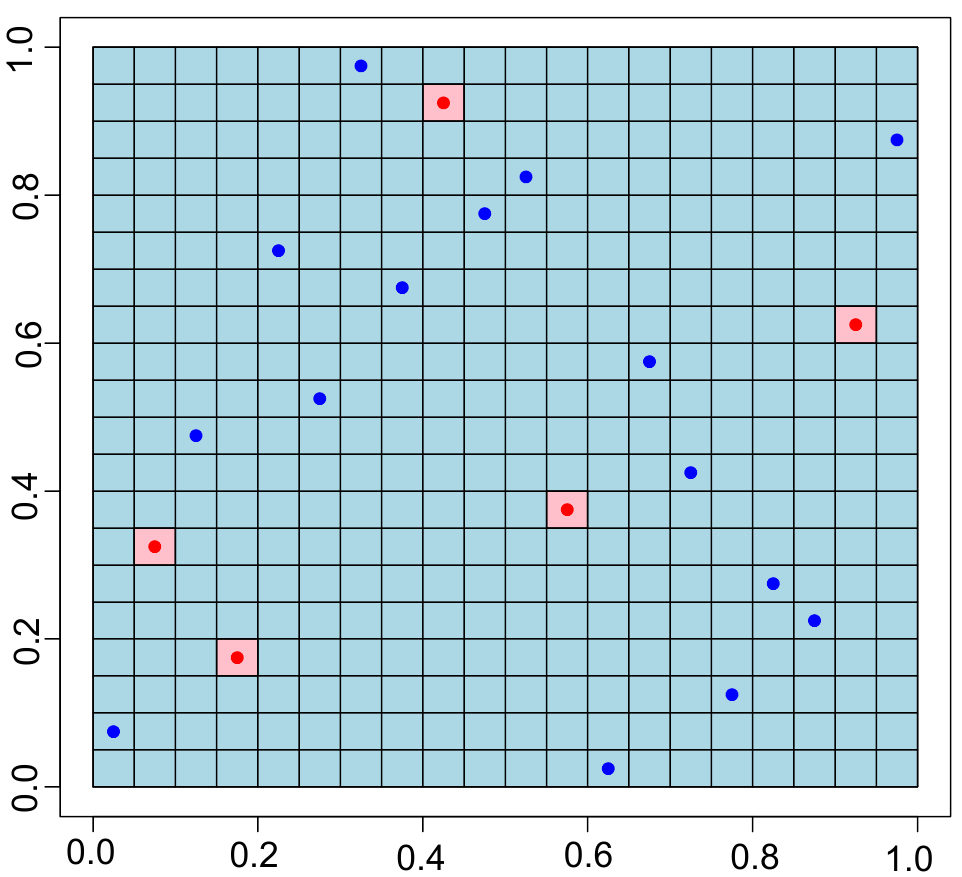}
        \caption{Hypothetical Classifier 1}
	\label{subfig:hc1}
     \end{subfigure}
     \quad
     \begin{subfigure}[b]{0.48\textwidth}
        \includegraphics[width=\textwidth]{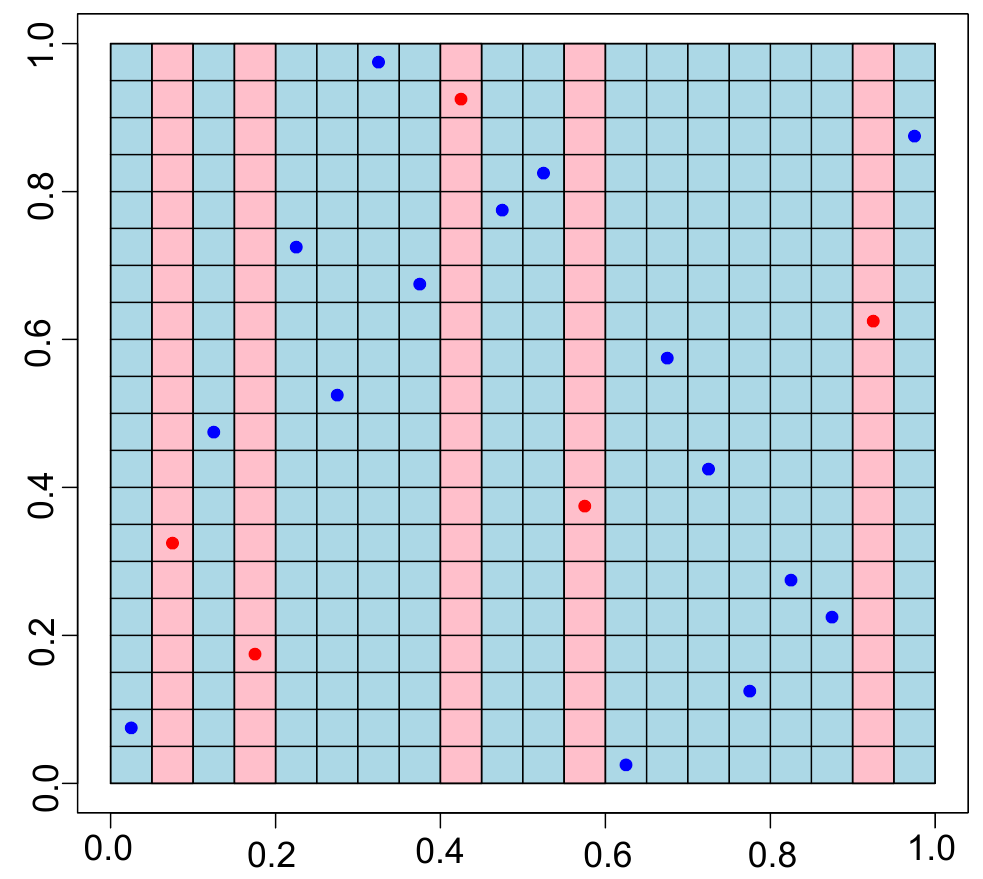}
        \caption{Hypothetical Classifier 2}
	\label{subfig:hc2}
     \end{subfigure}
     \begin{subfigure}[b]{0.48\textwidth}
        \includegraphics[width=\textwidth]{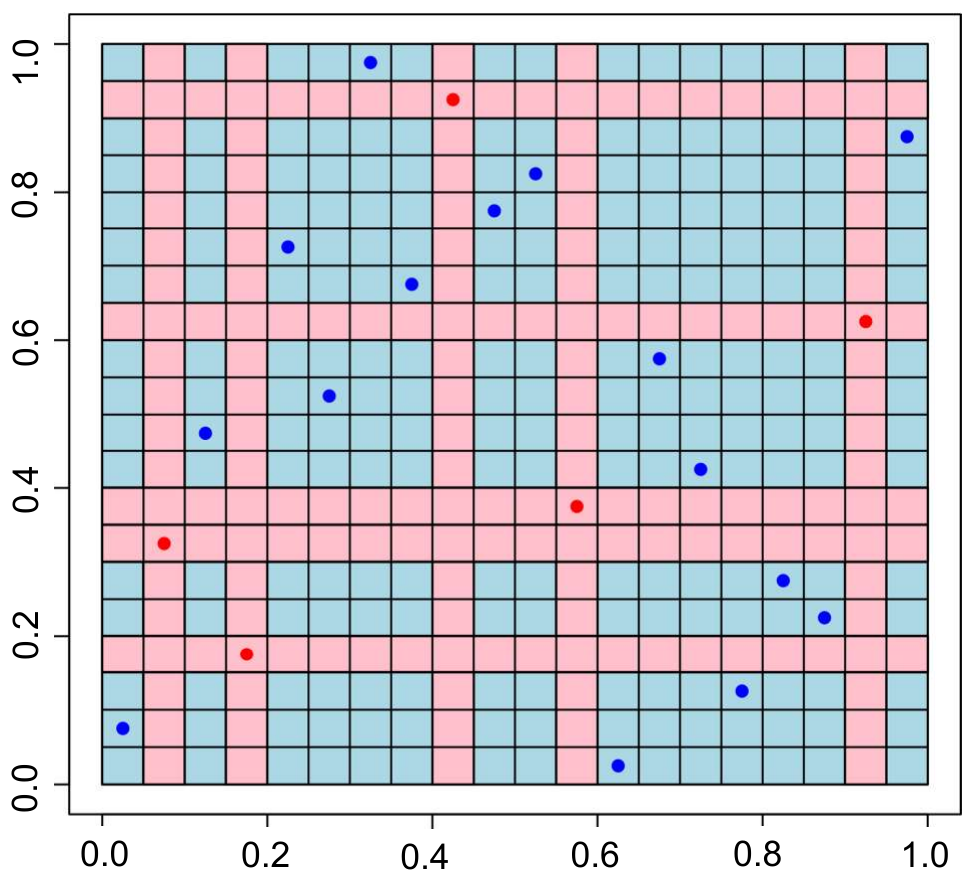}
        \caption{Hypothetical Classifier 3}
	\label{subfig:hc3}
     \end{subfigure}
     \begin{subfigure}[b]{0.48\textwidth}
        \includegraphics[width=\textwidth]{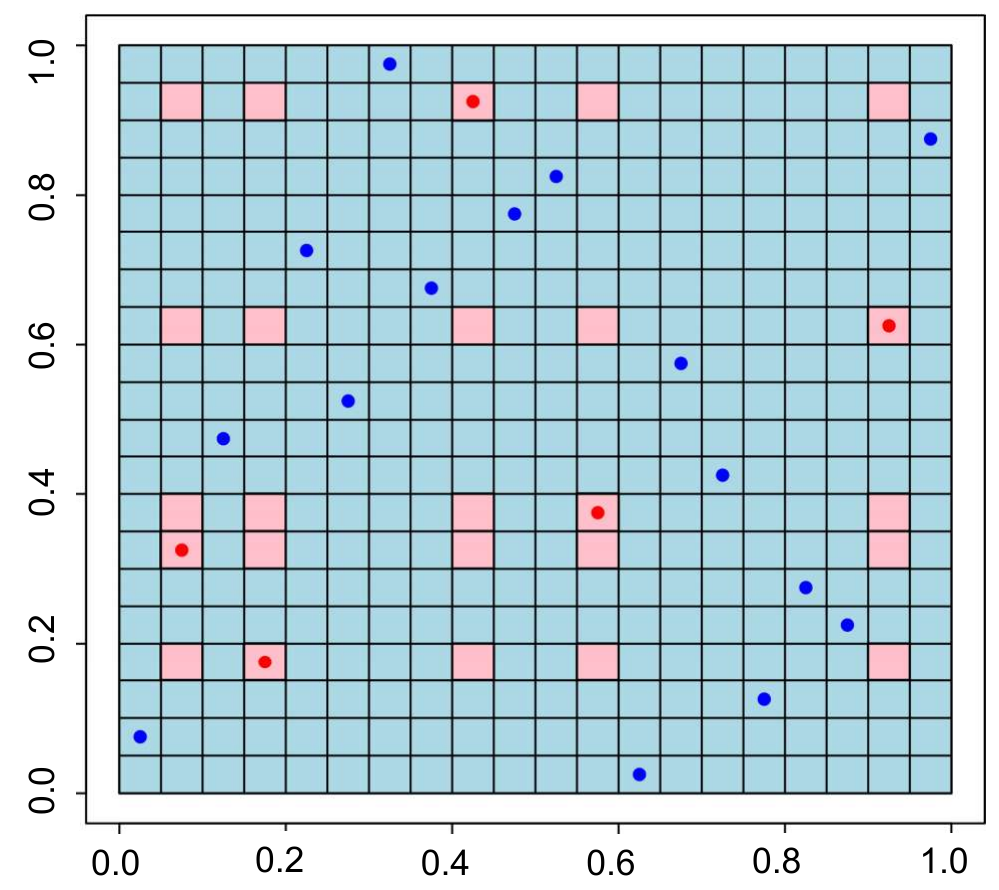}
        \caption{Hypothetical Classifier 4}
	\label{subfig:hc4}
     \end{subfigure}
      \caption{Four different hypothetical classifiers on a pure noise response surface where $\cpr = 0.75$.} \label{fig:2d}
	\label{fig:four_hcs}
\end{figure}

 Figure~\ref{fig:four_hcs} shows four hypothetical classifiers that could result from fitting boosted decision tree models to the training data in Figure~\ref{fig:trainingdata}.
When decision trees are used as base learners for data with continuous predictors, it is a common convention to
restrict the split points of the trees to be the midpoints of the
predictors in the training data.  Consequently,  the classifier in each small square
shown in Figure \ref{fig:trainingdata} will necessarily be constant throughout; this is the finest resolution of the
classifier resulting from boosting decision tress assuming no
sub-sampling. Thus, Figure~\ref{subfig:hc1}
represents the  interpolating classifier closest to the Bayes
rule.  (In these plots, pink squares represent ``-1"'s and light
blue squares represent ``+1'''s.) Note that the interpolation is in fact quite local; the estimated function varies rapidly in the small neighborhoods of the pink squares.  For such a classification rule
the percentage of points in a hold-out sample that would differ
from the Bayes rule (in expectation over training sets) would be $(1-p)n/n^d$ where
$p=\cpr$ and $d$ is the dimensionality of the predictor space (for our example, $d=2$).  We
will present evidence later that in noisy environments boosting
sufficiently large trees does actually tend to find such rules as that in Figure~\ref{subfig:hc1}.
Interestingly, since
\[ \lim_{n \rightarrow \infty} (1-p)n/n^d =0 \] such rules are in fact
consistent.  This illuminates the point that interpolation does
not rule out consistency.  By allowing the decision boundary to be ``mostly" smooth, with spikes of vanishing measure, it is possible to obtain consistency  in the limit as $n \rightarrow \infty$,  even while classifying every training point correctly. This stands in direct  contrast to
the conclusion of others such as \citet{Bickel2006} who have observed that the ``empirical optimization problem
necessarily led to rules which would classify every training set
observation correctly and hence not approach the Bayes rule
whatever be $n$." 

While a classifier such as that in Figure~\ref{subfig:hc1} would preform well and is even consistent, many
possible interpolators exist, such as the others displayed in
Figure~\ref{fig:four_hcs}.  Figure~\ref{subfig:hc2} shows the (hypothetical) result of
allowing the boosting algorithm to use trees involving only $x_1$
and not $x_2$.
It is interesting to note that this classifier has severely overfit, even though it is a simpler model,  depending  on only one of the two predictors.  The classifier in Figure~\ref{subfig:hc3} has an even worse error rate, while the classifier in Figure~\ref{subfig:hc4} differs from the Bayes rule with rate
$((1-p)n)^2/n^2$.  This final example illustrates the type of
structure and error rate that  occurs when stumps are used
as the weak learner. In fact, \citet{Mease2008} show that the
additive nature of stumps results in  boosted classifiers that differs
from the Bayes rule at a rate of at least $(1-p)^d(1-1/d)^d$ and
hence is not consistent.  The reason for this is that using linear
combinations of stumps does not provide enough flexibility to
interpolate  locally around points for which the observed class
differs from the Bayes rule. In contrast, boosting  larger trees, such as those grown in random forests interpolating with spikes of increasingly smaller size. Some simulations demonstrating the superior performance of larger
trees over stumps are given in \citet{Mease2008} and here in Section~\ref{subsec:stumps}.

The different classification rules represented by
the four plots all interpolate the training data; however, their
performances on the population vary considerably due to 
different degrees of local interpolations of noise. In the sequel, we 
will show how random forests and boosted ensembles of large trees results in classifiers that are
robust to noise. The classifiers behave in noisy regions as in Figure~\ref{subfig:hc1}.
AdaBoost and random forests  average many individually overfit classifiers, similar to the one in Figure~\ref{subfig:hc2}.  The result is a final
robust classifier, that is spiked-smooth; it fits the noise  but only extremely locally.

\subsection{A Two-Dimensional Example with Pure Noise}
\label{subsec:2d_noise}

We will begin with an easy to visualize example that demonstrates how fine interpolation can provide robustness in a noisy setting.  In particular, we compare the performance of AdaBoost,  random forests and one-nearest neighbors,  which are all interpolating classifiers.  We will see see graphically that AdaBoost and random forests interpolate more locally around error points in the training data than the one-NN classifier.  Consequently, AdaBoost and random forests are less  affected by noise points as one-NN and have lower generalization error.  We will show that the self-averaging property of AdaBoost and random forests is  crucial. This property will be discussed in subsequent sections.

The implementation of AdaBoost is carried out according to the
algorithm described earlier.  The base learners used are trees fit
by the \texttt{rpart} package \citep{Therneau1997} in \texttt{R}.  The trees are grown to a maximum
depth of 8, meaning they may have at most $2^8=256$ terminal
nodes. This will be the implementation of AdaBoost we will
consider throughout the remainder of this paper.

We will consider again the ``pure noise'' model as described in
the previous section,  where  the probability that 
$y$ is equal to $+1$  for every $\x$ is some constant value $p>.5$.  For the training
data we will take $n=400$ points uniformly sampled on $[0,1]^2$
according to a Latin Hypercube using the midpoints as before. For
the corresponding $y$ values in training data we will randomly
choose 80  points to  be $-1$'s so that $\cpr=.8$.

Figure \ref{fig:2d} displays the results for the following:   (a) one-NN, (b) AdaBoost , and (c) random forests.  Regions classified as $+1$ are colored light blue and regions classified as $-1$ are colored
pink. The training data is displayed  with
blue points for $y=+1$ and red points for $y=-1$.  Since the
Bayes' rule would be to classify every point as $+1$, we judge the 
performance of the classifiers by the fraction of
the unit square that matches the Bayes' rule.  The nearest neighbor
rule in this example classifies $79\%$ of the region as $+1$ (we expect $p=80\%$ on average for the one-NN) while AdaBoost
performs substantially better classifying $87\%$ of the square as
$+1$ after 100 iterations (which is long after the training error equals 
zero).  This is evidence of boosting's robustness to noise
discussed in the previous section.   The random forests (with 500 trees) does even better, classifying $94\%$ of the figure as $+1$.  Visually, it is obvious that the random forests and AdaBoost classifier is more spiked-smooth than one-nearest neighbors, which allows it to be less sensitive to noise points.  AdaBoost and random forests do in fact overfit the noise---but only the noise.  They do not allow the overfit to metastasize to modestly larger neighborhoods around the errors.  It is interesting to note that there seems to be a large degree of overlap between the regions classified as -1 by both the random forests and AdaBoost; one-NN does not seem to visually follow a similar pattern. 

 As we will see in the Section~\ref{subsec:twenty_dim}, by increasing the sample size, number of dimensions and iterations the performance is even better.  The agreement with the Bayes rule for AdaBoost and random forests converge to practically $100\%$ despite the fact that both algorithms still interpolate the training data without error.

\begin{figure}
\centering
    \begin{subfigure}[b]{0.4\textwidth}
        \includegraphics[width=\textwidth]{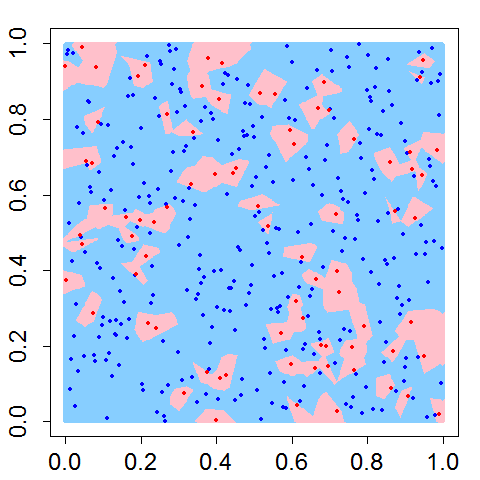}
        \caption{one-NN}
     \end{subfigure}
     \quad
     \begin{subfigure}[b]{0.4\textwidth}
        \includegraphics[width=\textwidth]{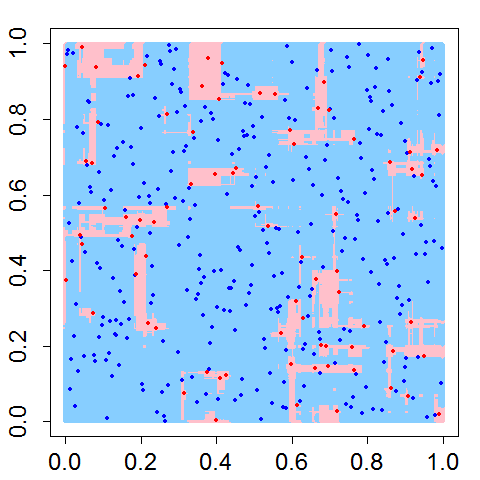}
        \caption{AdaBoost}
     \end{subfigure}
     
     \begin{subfigure}[b]{0.4\textwidth}
        \includegraphics[width=\textwidth]{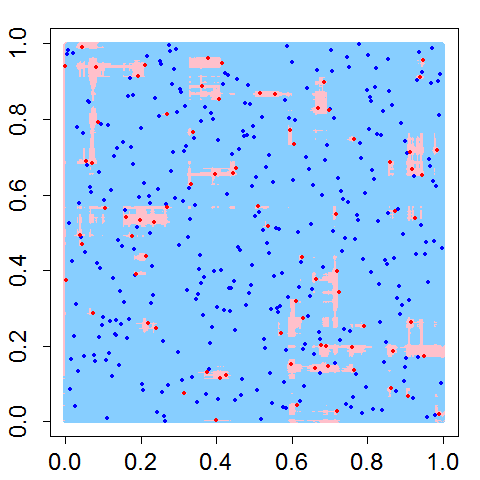}
        \caption{Random Forests}
     \end{subfigure}
      \caption{Performance of one-NN, AdaBoost, and random forests on a pure noise response surface with $\cpr = .8$ and  $n=400$ training points. } \label{fig:2d}
\end{figure}

\subsection{A Visualization of Spiked-Smoothing} \label{subsec:spikedsmoothing}

\begin{figure}[htp]
\centering
    \begin{subfigure}[b]{0.3\textwidth}
        \includegraphics[width=\textwidth]{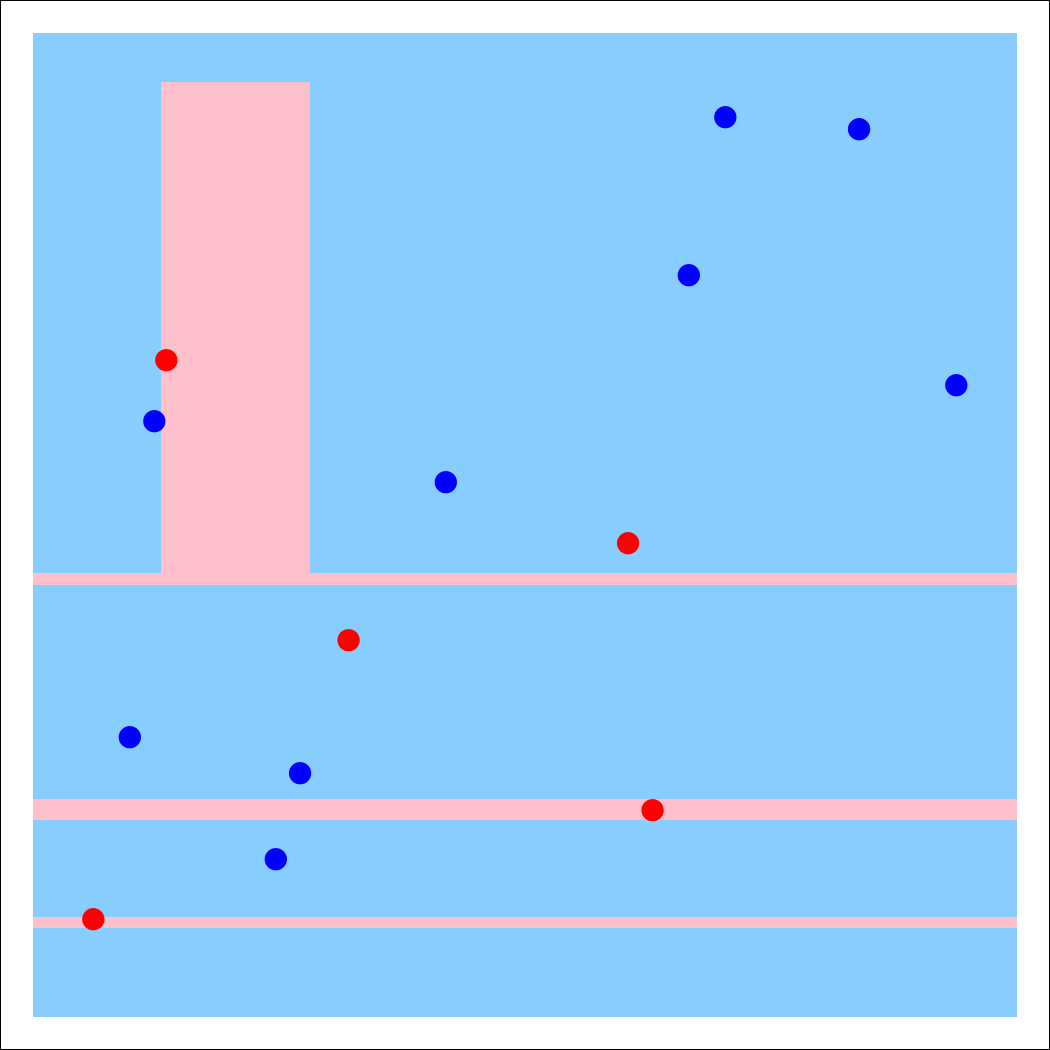}
        \caption{Tree 1}
	\label{subfig:rf1}
     \end{subfigure}
     \quad
     \begin{subfigure}[b]{0.3\textwidth}
        \includegraphics[width=\textwidth]{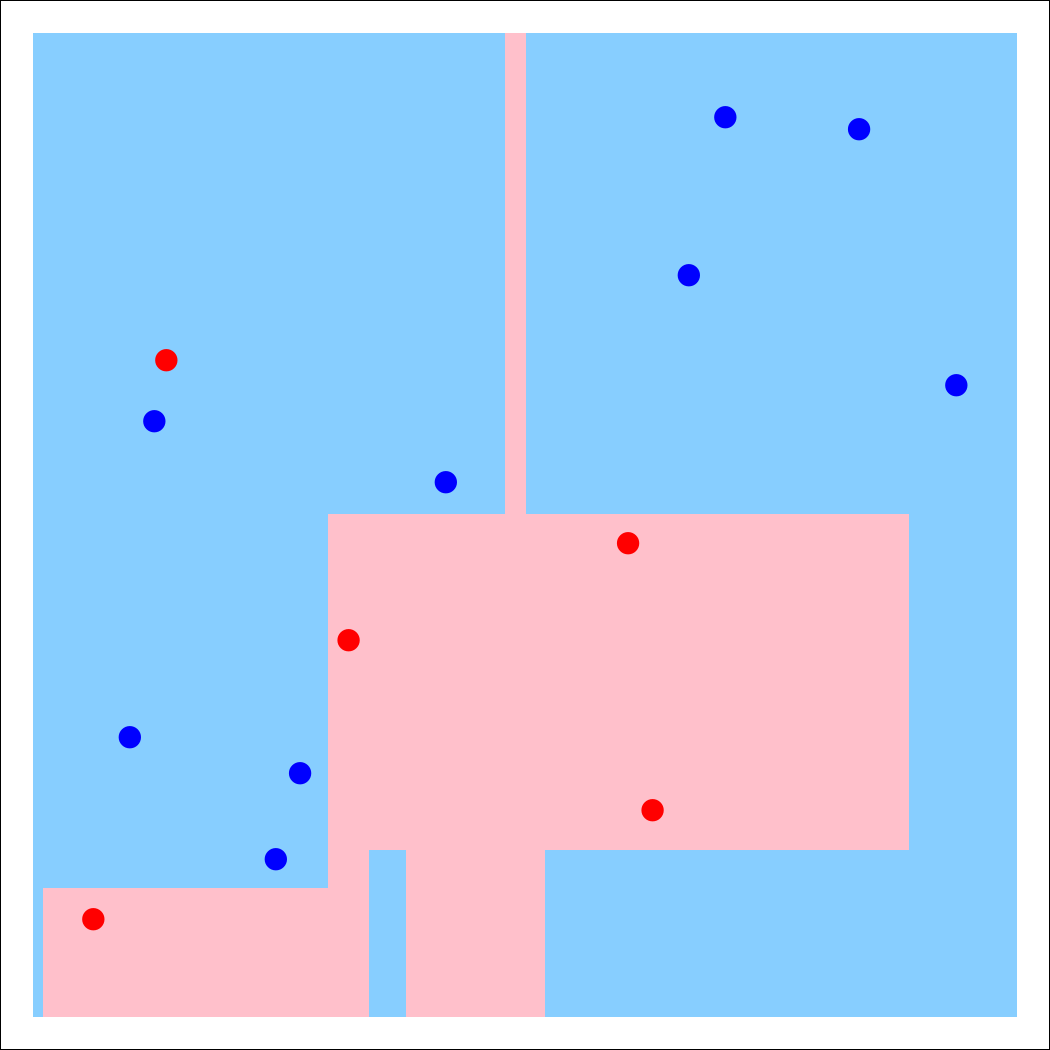}
        \caption{Tree 2}
	\label{subfig:rf2}
     \end{subfigure}
     \quad
     \begin{subfigure}[b]{0.3\textwidth}
        \includegraphics[width=\textwidth]{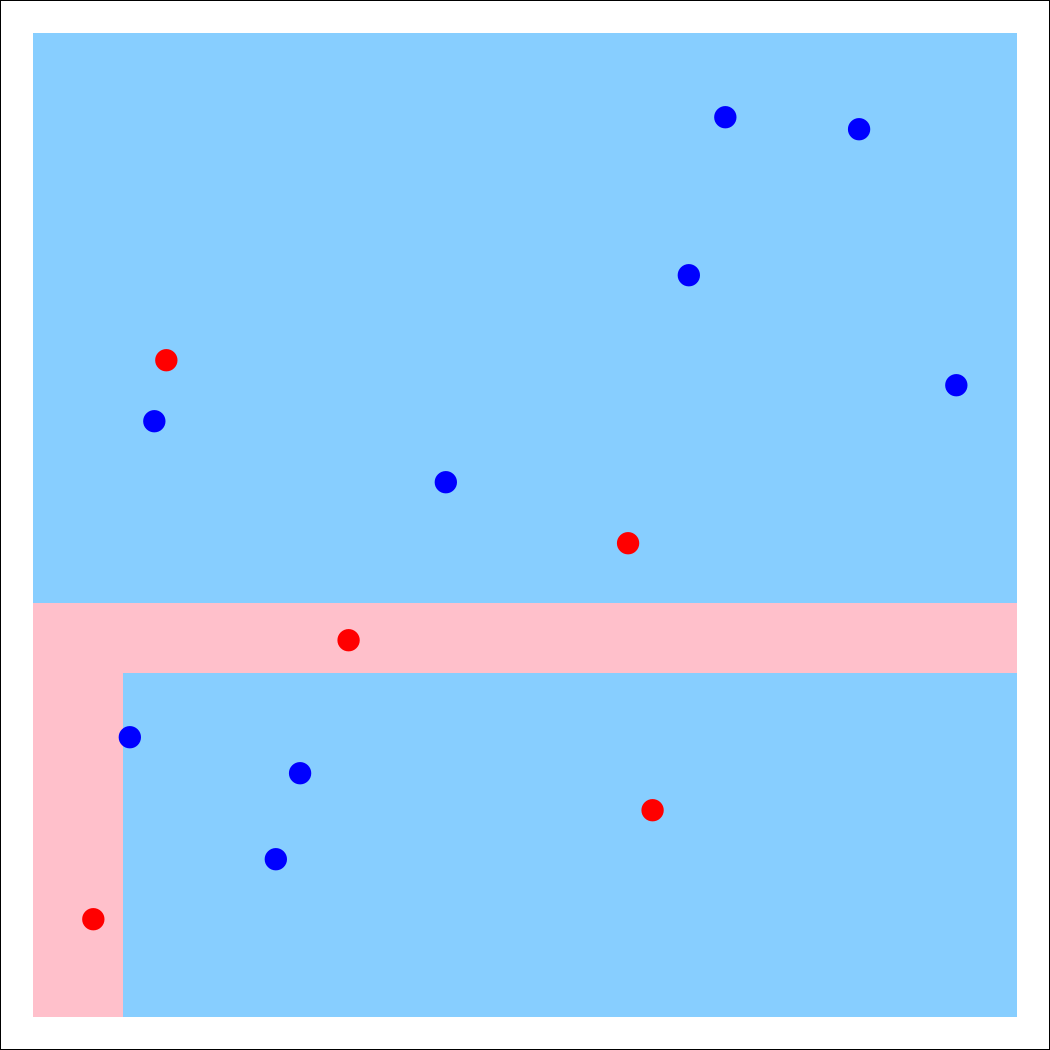}
        \caption{Tree 3}
	\label{subfig:rf3}
     \end{subfigure}
     \quad 
     \begin{subfigure}[b]{0.3\textwidth}
        \includegraphics[width=\textwidth]{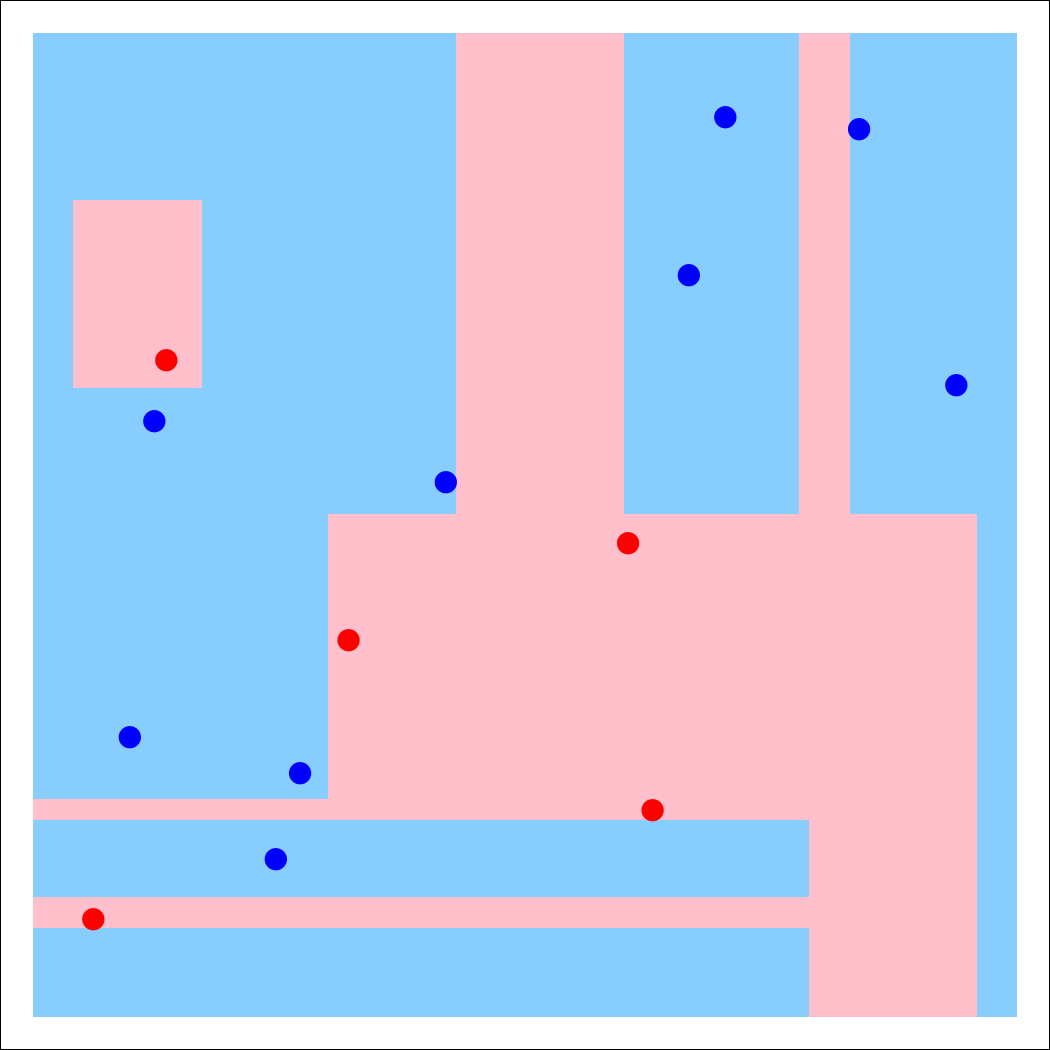}
        \caption{Tree 4}
	\label{subfig:rf4}
     \end{subfigure}
     \quad
         \begin{subfigure}[b]{0.3\textwidth}
        \includegraphics[width=\textwidth]{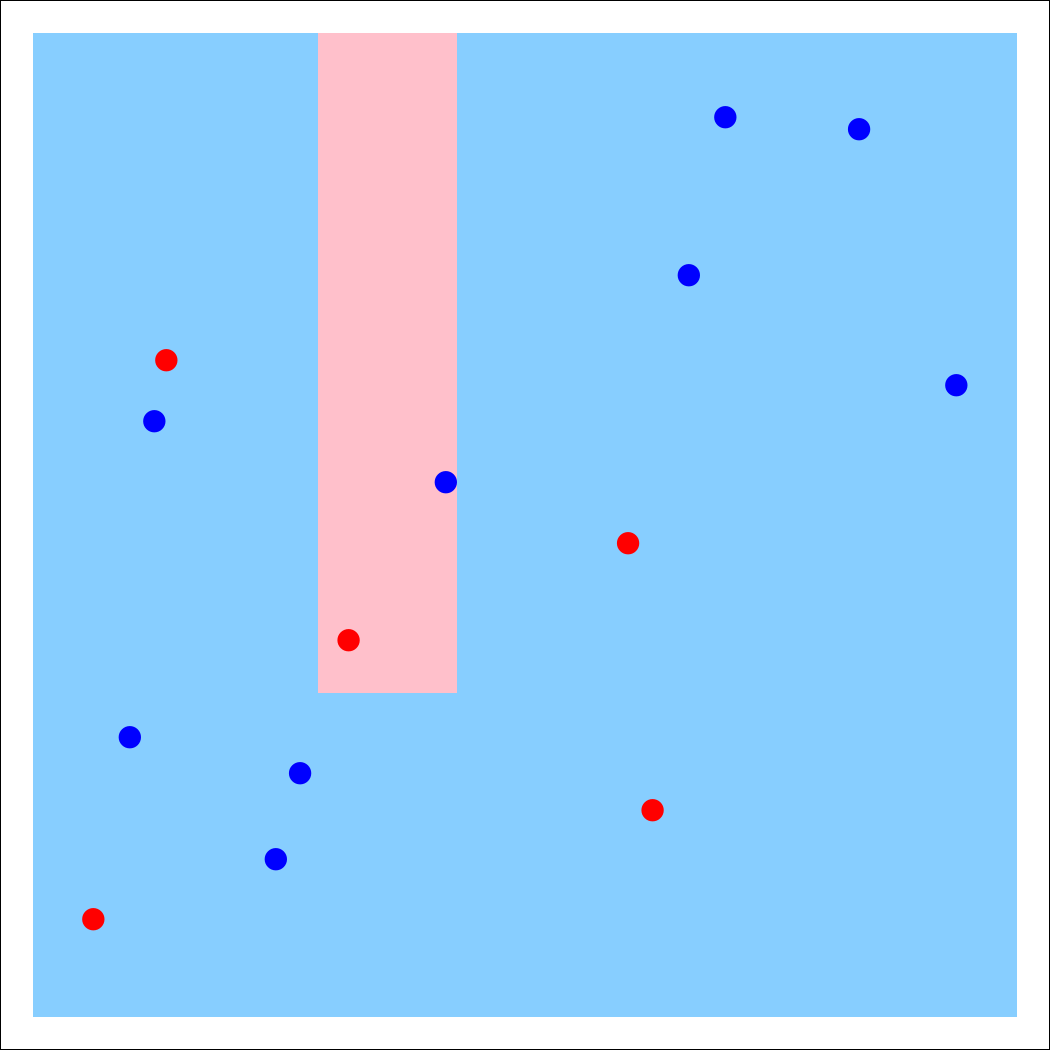}
        \caption{Tree 5}
	\label{subfig:rf5}
     \end{subfigure}
     \quad
         \begin{subfigure}[b]{0.3\textwidth}
        \includegraphics[width=\textwidth]{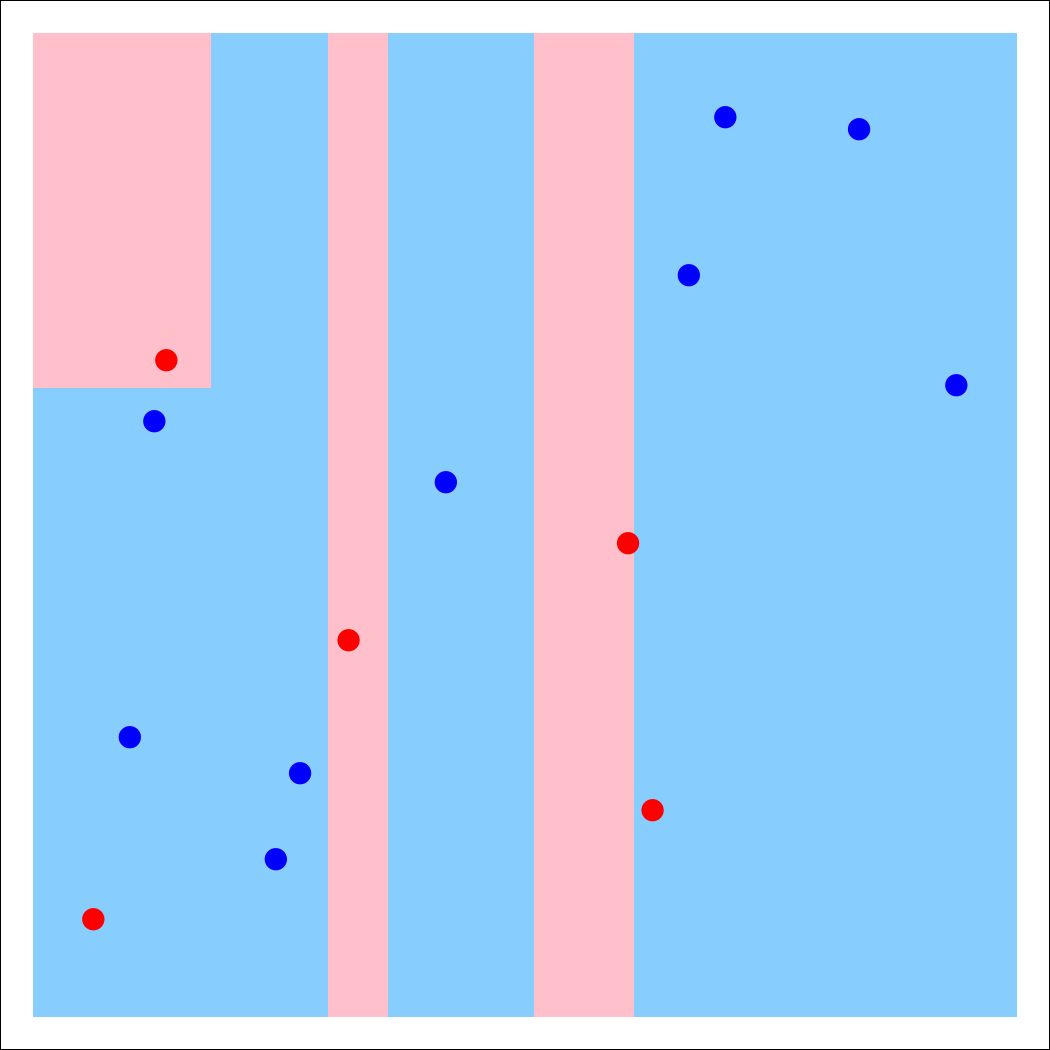}
        \caption{Tree 6}
	\label{subfig:rf6}
     \end{subfigure}
	         \begin{subfigure}[b]{0.3\textwidth}
        \includegraphics[width=\textwidth]{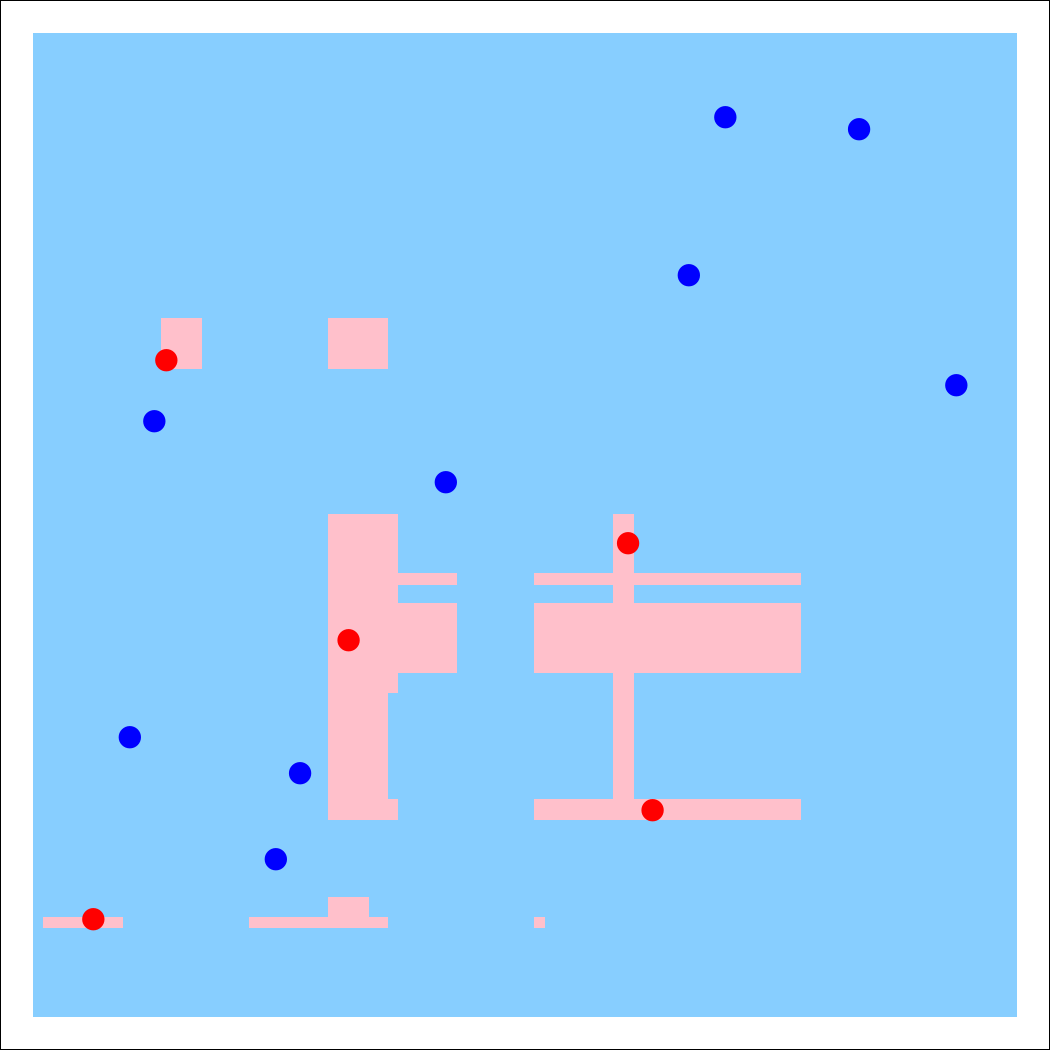}
        \caption{Majority Tree Vote}
	\label{subfig:rfvote}
     \end{subfigure}
      \caption{First six trees from a random forest, along with the classifier created by a majority vote over the trees.} \label{fig:rfDecomp}
\end{figure}

We have argued that local interpolation such as in Figure~\ref{subfig:hc3} is desirable, and we have demonstrated that AdaBoost and random forest classifiers can achieve such a fit in the previous simulation.  Now, we turn to the crucial point of how these classifiers achieve such a fit.  To this end, we will graphically display in the process of spiked-smoothing in the case of the random forest classifier from the previous simulation.  Each of the first six plots in Figure~\ref{fig:rfDecomp} shows the classification rule fit by different decision trees in the random forest.  We have restricted each plot to a subset of the unit square to aide in visual ease.  The bottom plot, Figure~\ref{subfig:rfvote} shows the classifier created from a majority vote of each of the six random forest decision trees.  As in the previous sections, the light blue regions indicate where a classifier returns $y=+1$, and the pink regions indicate where a classifier returns $y=-1$.

	   As before, we remark that the Bayes rule in this case would be to classify every point as $y=+1$, and so agreement with the Bayes rule in the plots below can be visualized as the proportion of the figure that is light blue.  The first thing to notice is that each decision tree fails to reproduce the Bayes rule.  Indeed, since each tree interpolates its bootstrap sample, each figure is bound to contain regions of pink, since most bootstrap samples will contain at least a few noise points.  However, one will also notice that these regions of pink tend to be localized into thin strips (this is especially apparent in trees one, three, five, and six).  In other words, noise points tend not to ruin the fit of the decision tree at nearby points..  The magic of spiked-smoothing is revealed in the classifier ~\ref{subfig:rfvote} created by a majority vote of the six decision trees.  By itself, each decision tree is a poor classifier (evinced by relatively large regions of pink).  However, when voted these regions of pink get shrunk down into smaller regions, indicating better agreement with the Bayes rule.  One can easily imagine that if these ``thin strips" were actually much wider, as in the case of fitting stumps, averaging would not be able to reduce the influence of these noise points enough.  The end effect of averaging is to create a decision surface which is affected only very minimally by the noise points in the training set.  A simulation in Section~\ref{sec:avg_boost} will demonstrate that the additional iterations of AdaBoost serve to ``shrink" the fit around noise points, much as the regions of pink in this example became more localized after averaging.

\subsection{A Two-Dimensional Example with Signal}

In light of the example in the previous section, one might note that certain non-interpolating algorithms, such as a pruned CART tree, would recover the Bayes error rate exactly. In this section, we consider an example where a much more complex classifier is required to recover the signal, yet the self-averaging property is still needed to prevent over-fitting to noise.

We consider $n=1000$ training points sampled uniformly on $[0,1]^2$ with the Latin Hypercube design. In this simulation, there is signal present.   Inside of a circle of radius $0.4$ centered in the square, the probability that $y=+1$ is set to $0.1$, while the probability that $y=+1$ outside the circle is set to $0.9$.

This simulation setting is similar to the previous one, except that the probability that $y=+1$ varies at different points over the unit square.  One can see in Figure \ref{fig:2dNoise} that the Bayes rule in this setting is just to label every point inside the circle $y=+1$ and every point outside the circle $y=-1$, which gives a Bayes error rate of $0.1$. We can then compare the performance of AdaBoost, random forests, and CART as in the previous section by examining how much of the circle gets classified as $y=+1$ and how much of the outer region is classified as $y=-1$. We run AdaBoost for 500 iterations, fit a random forests model with 500 trees, and build a CART tree that is pruned via cross-validation. Note that we prune the CART tree in order to show how a ``classical'' statistical model of limited complexity performs on the classification task.  

We find that AdaBoost and random forests have an overall error rate of around $0.13$, one-nearest neighbor has an overall error rate of $0.20$, and CART has an error rate of $0.18$.  CART fails to perform well in this example because it is not allowed enough complexity to capture the circular pattern. To do so via only the splits parallel to the axes allowed by the algorithm would require a very deep tree (as allowed in random forests and AdaBoost), which pruning does not afford. Rather, a shallow tree can only recover a simple rectangular pattern due to its shallow depth. One-NN, on the other hand, again suffers from its inability to keep the interpolation localized. Outside of the circle, one can observe small ``islands" of pink surrounding noise points: by failing to localize the fit, test points near these noise points get classified incorrectly. Again, one finds that random forests and AdaBoost have superior performance because they tend to finely interpolate the training data, and the process of spiked-smoothing shrinks down the influence of noise points. 

\begin{figure} \label{fig:2dNoise}
        \centering
        \begin{subfigure}[b]{0.48\textwidth}
                \includegraphics[width=\textwidth]{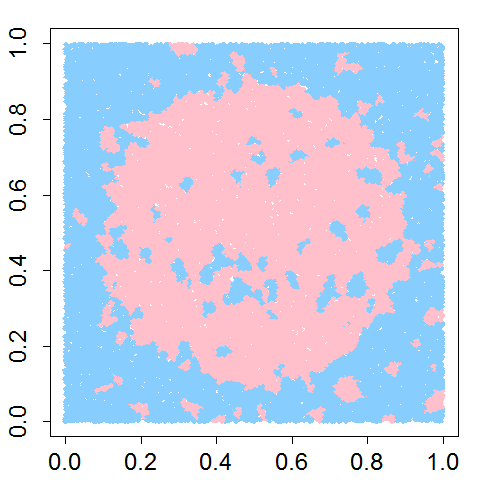}
                \caption{One-NN}
        \end{subfigure}
        \quad
        \begin{subfigure}[b]{0.48\textwidth}
                \includegraphics[width=\textwidth]{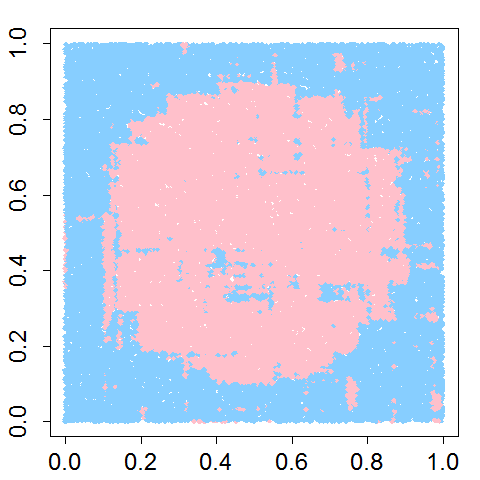}
                \caption{AdaBoost}
        \end{subfigure}

        \begin{subfigure}[b]{0.48\textwidth}
                \includegraphics[width=\textwidth]{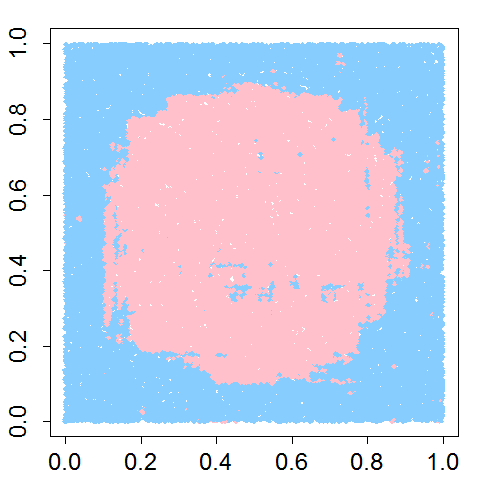}
                \caption{Random Forests}
        \end{subfigure}
        \quad
         \begin{subfigure}[b]{0.48\textwidth}
                \includegraphics[width=\textwidth]{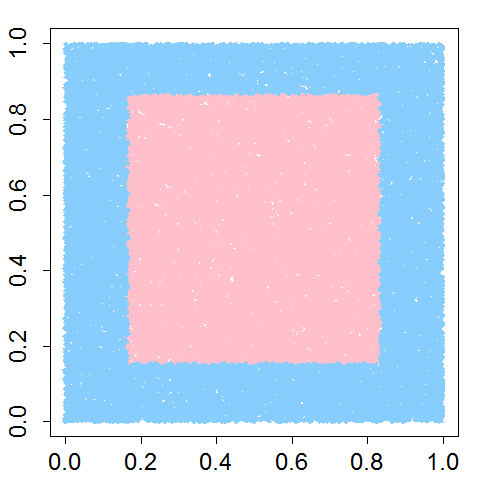}
                \caption{CART}
        \end{subfigure}
        \caption{Performance of AdaBoost, random forests, and CART on a response surface where $\cpr = 0.10$ inside the circle and $\cpr)=0.90$ outside of the circle. There are $n=1000$ training points and the Bayes error is 0.10.}
\label{fig:2dNoise}
\end{figure}

\subsection{A Twenty-Dimensional Example}\label{subsec:twenty_dim}

We now repeat the simulation in Section~\ref{subsec:2d_noise} with a larger
sample size and in 20 dimensions instead of 2.  Specifically, the
training data now has $n=5000$ observations sampled according to
the midpoints of a Latin Hypercube design uniformly on
$[0,1]^{20}$. We again randomly select $20\%$ or 1000 of these
points to be $-1$'s with the remaining 4000 to be $+1$'s.

\begin{figure}[] 
 \centerline{\includegraphics[width=2.5 in]{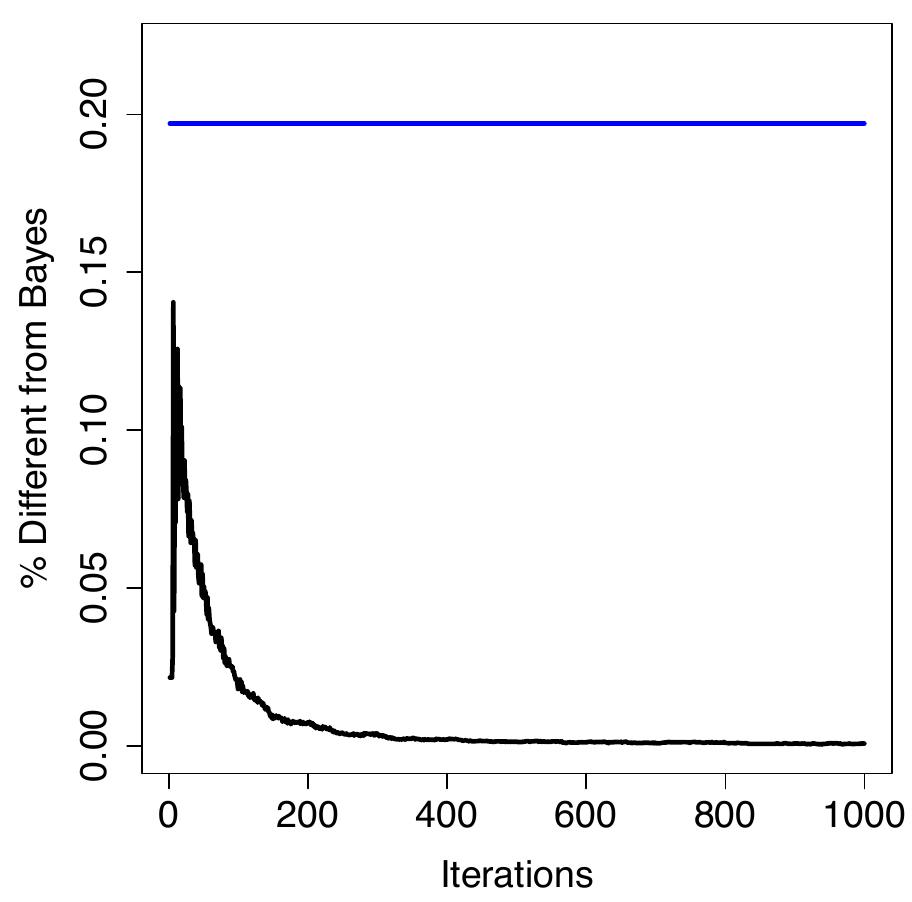}}
 \caption{This plot shows the proportion of points in a test set 
 for which the predictions made by AdaBoost and the Bayes rule
 differ, as a function of the number of boosting rounds (black).
 The blue line shows this proportion for the one nearest neighbor
 classifier.  Note that the agreement of AdaBoost and the Bayes rule increases with the number of boosting rounds.}
  \label{fig:20dplot}
\end{figure}

Since in 20 dimensions it is difficult to display the resulting
classification rules graphically we instead examine the rules on a
hold out sample of 10,000 points sampled uniformly and
independently on $[0,1]^{20}$.  Figure~\ref{fig:20dplot} plots the
proportion of points in the hold out sample classified by
AdaBoost as $+1$ as a function of the number of iterations.  This
proportion peaks at $.1433$ at nine iterations but then gradually
decreases to $.0175$ by 100 iterations and is equal to $.0008$ by
1,000 iterations. The fact that by 1,000 iterations only 8 of the
10,000 points in the hold out sample are classified as $+1$
means there is very little overfitting.  The large number of
iterations has the effect of smoothing out the classifier
resulting in a rule that agrees with the Bayes rule for $99.92\%$
of the points.  Recall that AdaBoost fits the training data perfectly, and thus
differs from the Bayes rule on $20\%$ of this sample.  We see clearly here that AdaBoost overfits with respect to the training data but not with respect to
the population.  Again, this is a result of extremely local
interpolation of the points in the training data for which the
observed class differs from the Bayes rule.  A random forests model fit to the training data agrees with the Bayes rule at every point except for one, and hence
has exceptional generalization error.

\section{Self-Averaging Property of Boosting}\label{sec:avg_boost}

\subsection{Boosting is Self-Smoothing}

In the previous sections, we have demonstrated simple examples where random forests and AdaBoost yield the strongest performance with respect to the Bayes rule. We have argued that these algorithms are successful classifiers due to the fact that they fit initially complex models by interpolating the training data but also exhibit smoothing properties via self-averaging that stabilizes the fit in regions with signal, while continuing to keep localized the effect of noise points on the overall fit. While this smoothing mechanism is obvious for random forests via the averaging over decision trees, it is less obvious for AdaBoost. In this section we explain why the additional iterations in boosting way beyond the point at which perfect classification of the training data (i.e interpolation) has occurred actually has the effect of smoothing out the effects of noise rather than leading to more and more overfitting. To the best of our knowledge, this is a novel perspective on the algorithm. To explain our key idea, we will recall the pure noise example from before with $p=.8$, $d=20$ and $n=5000$.

Recall that the classifier produced by AdaBoost corresponds to
$\I[f_M(x)>0]$ where

\begin{equation*}
f_M(x)=\sum_{m=1}^{M} \alpha_m G_m(x)   
\end{equation*}

as defined earlier.  Taking $M=1000$ which was successful in our
example let us rewrite this as

\begin{equation*}
f_{1000}(x)=\sum_{m=1}^{1000} \alpha_m G_m(x)=\sum_{j=1}^{10}\sum_{k=1}^{100} \alpha_{100(j-1)+k} G_{100(j-1)+k}(x)= \sum_{j=1}^{10}\sum_{k=1}^{100} h_k^j(x)   
\end{equation*}

where
\begin{equation*}
 h_k^j(x) \equiv  \alpha_{100(j-1)+k} G_{100(j-1)+k}(x).
\end{equation*}

\begin{figure}[bhtp]
 \centerline{\includegraphics[width=7in]{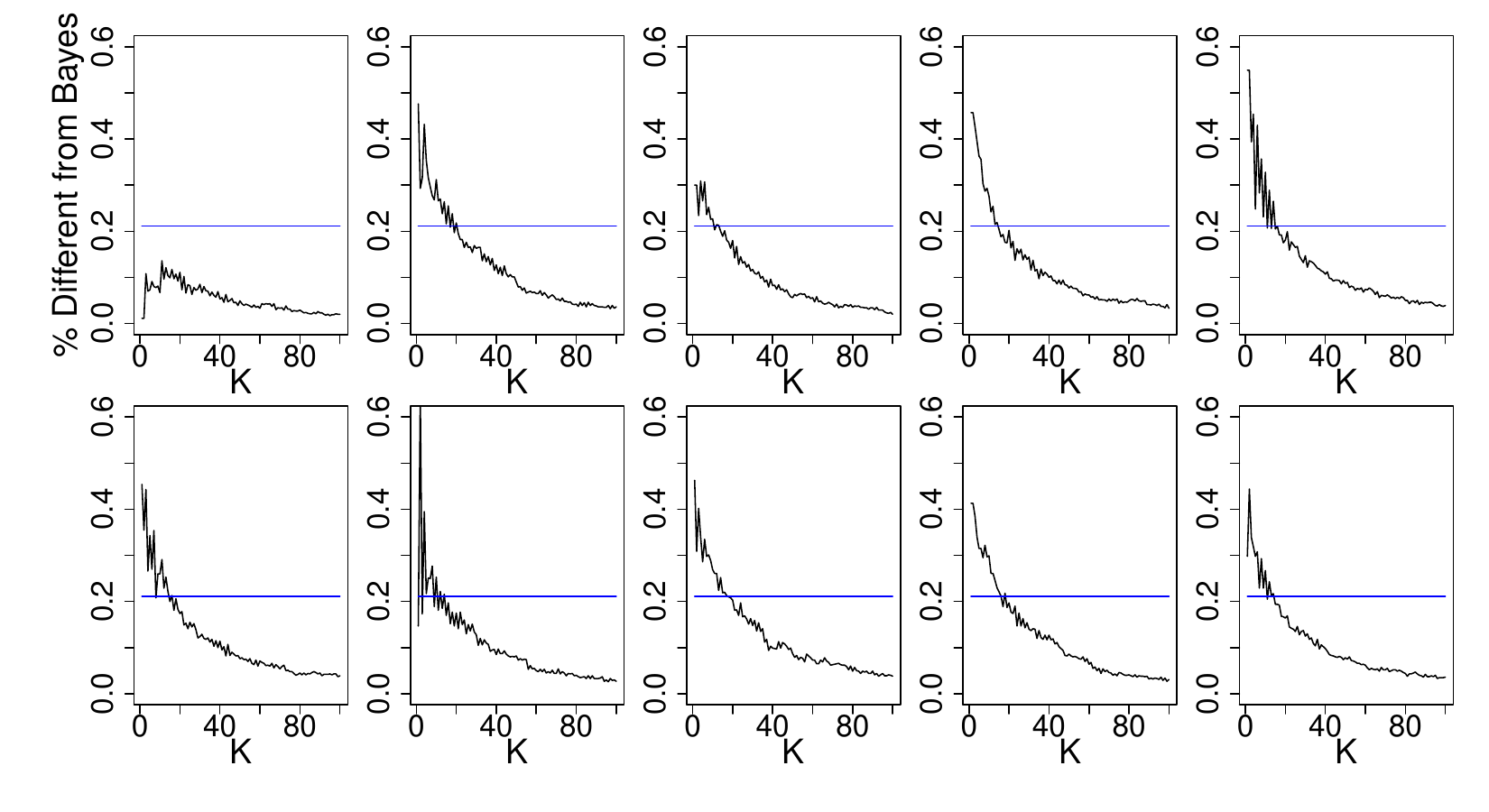}}
 \caption{A decomposition of boosting} \label{fig:selfsmooth}
\end{figure}

Now define

\begin{equation*}
 h_K^j(x) \equiv \sum_{k=1}^{K} h_k^j(x) 
\end{equation*}

and note that for every $j \in \{1,...,10\}$ and every $K \in
\{1,...100\}$ that $\I[h_K^j(x)>0]$ is itself a classifier made by
linear combinations of classification trees.  The ten plots in
Figure \ref{fig:selfsmooth} display the performance on the hold-out
sample for these ten classifiers corresponding to the ten
different values for $j$ as a function of $K$.  Interestingly,
each of these 10 classifiers by itself displays the characteristic
of boosting the agreement with the Bayes rule increases as
more terms are added (for instance, as $K$ is increased).

A second interesting fact about these 10 individual classifiers in
the decomposition is that each one achieves perfect separation of
the training data and thus each one is an interpolating
classifier.  This result can be expected in general, provided the
total number of iterations for each classifier in the
decomposition is sufficiently large.  This is clear for the first
classifier, since it is simply AdaBoost itself and will
necessarily achieve zero training error under some standard
conditions as discussed in \citet{Jiang2002}.  The second classifier in
the decomposition is simply AdaBoost weight 
carried over from the first classifier.  Since re-weighting the
training data does not prevent AdaBoost from obtaining zero
training error, the second classifier also interpolated eventually, as does the third,
and so on.

Decomposing boosting in this way offers an explanation of why the additional iterations lead to robustness  and better performance in noisy environments rather than severe overfitting.  In this example, AdaBoost for 100 iterations is an interpolating classifier.  It makes some errors, mostly near the points in the training data for which the label differs from the Bayes rule, although these are localized.  Boosting for 1000 iterations is thus a point-wise weighted average of 10 interpolating classifiers. The random errors near the points in the training data for which the label differ from the Bayes' rule cancel out in the ensemble  average and become even more localized.  Of course, the final classifier is still an interpolating classifier as it is an average of 10 interpolating classifiers. In this way, boosting is  self-smoothing, self-averaging or self-bagging process that reduces overfitting as the number of iterations increase. The additional iterations provide averaging and smoothing---not overfitting. Empirically this is very similar  to random forests and provides evidence that both algorithms, which perform well in our examples, actually do so using the same mechanism.

\begin{figure}[htp]
 \centerline{\includegraphics[width=3.5in]{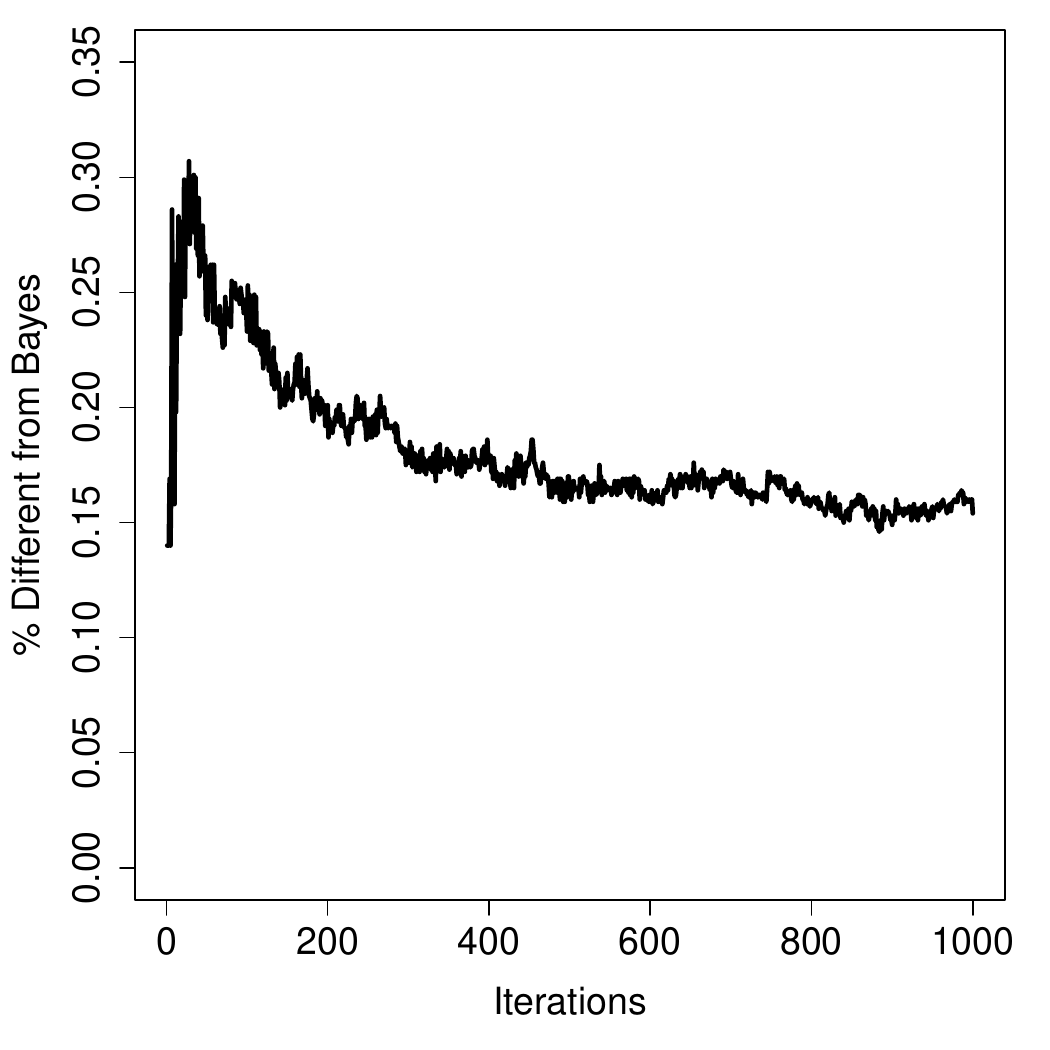}}
 \caption{This plot shows the proportion of points in a test set 
 for which the predictions made by AdaBoost and the Bayes rule
 differ, as a function of the number of boosting rounds (black).} 
\label{fig:shrink-neighbor}
\end{figure}

We further illustrate this phenomenon of increasing localization of the
interpolating resulting from this averaging through the following
simulation.  We take the same training data as before but this
time we form the hold out sample by taking a point a (Euclidean)
distance of .1 from each of the 1000 points labeled as $-1$ in
the training data in a random direction. Due to the  forced (and unnatural)
close proximity of the points in the hold out to training set deviations from the Bayes' rule (points with $-1$ labels), the error rate is much higher than it would be for a
random sample. However,  
the interpolation continues to become more localized as the
iterations proceed  (see Figure \ref{fig:shrink-neighbor}) so even points that are quite close to the label errors (the $-1$ points)
eventually become classified correctly as $+1$. Comparison to Figure
\ref{fig:20dplot} shows that this localization continues at a
steady rate even after the error on the random hold-out sample
is practically zero.  In contrast, the nearest neighbor
interpolator this simulation yielded $100\%$ disagreement with the
Bayes' rule.

\subsection{A Five-Dimensional Example}\label{subsec:five_dim}

We will now consider a second simulation to further illustrate how
this self-averaging property of AdaBoost helps prevent overfitting
and improves performance.  In this simulation we add 
signal while retaining significant random noise.  Let  $n=400$, $d=5$ and  sample $\x_i$ distributed
$iid$ uniform on $[0,1]^5$. The true model from for the 
simulation is
\[ \cpr=.2+.6 ~ ~ \Large{\I} \left [\sum_{j=1}^2 x_j>1 \right ]. \]
The Bayes' error is $0.20$ and the optimal Bayes' decision boundary is
the diagonal of the unit square in $x_1$ and $x_2$.  Even with this small sample size,   AdaBoost interpolates the training data after 10 iterations.  So we 
boost for 100 iterations  which decomposes into ten sets of
ten (which is analogous to the 10 sets of 100 from the 20 dimensional
example in the previous section).

\begin{figure}[bhtp]
 \centerline{\includegraphics[width=7in]{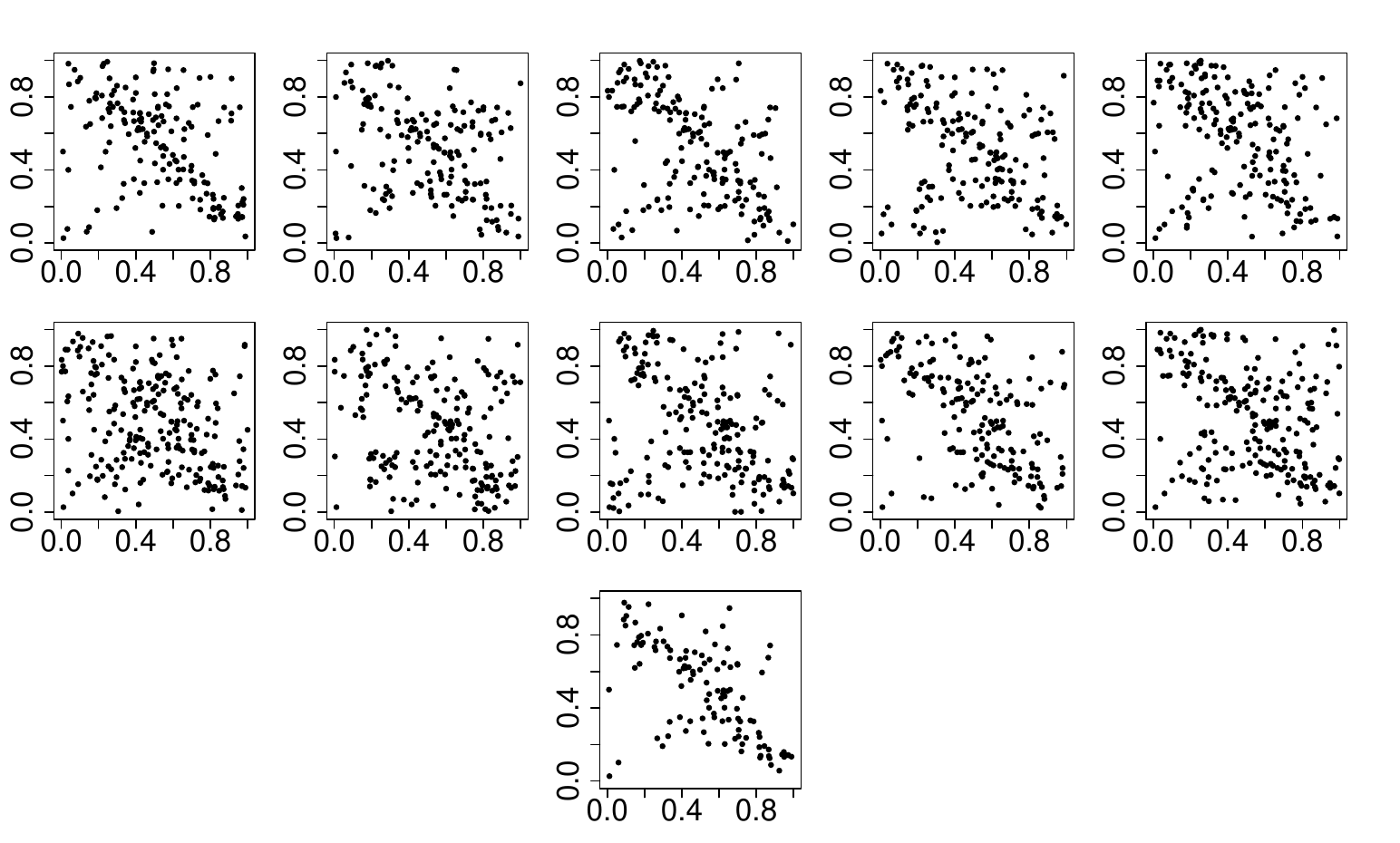}}
 \caption{Errors made by each classifier in a decomposition of boosting (first two rows) and errors made by the final classifier (bottom)} 
\label{fig:5d_example}
\end{figure}

The ten plots in the first two rows in Figure \ref{fig:5d_example} show the
performance of the ten classifiers corresponding to this
decomposition with respect to a hold out sample of 1000 points.
Each point in the figure represents a point classified differently
from the Bayes rule. While each of the ten classifiers in the
decomposition classifies a number of these 1000 points incorrectly
especially along the Bayes boundary, exactly which points are
classified incorrectly varies considerably from one classifier to
the next. The final classifier (displayed in the last plot), which
corresponds to AdaBoost after 100 iterations, makes fewer mistakes
than each of the ten individual classifiers. Since AdaBoost is a 
point-wise weighted average of the 10 classifiers,  the averaging
over the highly variable error locations made by each classifier reduces
substantially  the number of  errors made by the ensemble.  The percentage of
points classified differently from the Bayes' rule by the final classifier is 118/1000=0.118
while after the first ten there were still 162/1000=0.162
classified differently from the Bayes rule.  Averaged over 200 repetitions of this simulation
these numbers are .19 after the first ten iterations and .15 after
100 iterations confirming that the performance does improve by
running beyond the point at which interpolation initially occurs
as (a result of this self-smoothing).  

In this example, we also have evidence that AdaBoost takes some measure to decorrelate the errors made by its base classifiers, as hypothesized in \cite{Amit} and suggested in Figure ~\ref{fig:5d_example}.  To this end, we computed the correlation between $\mathbb{I}[h^1_{10}(X) = Y], \ldots, \mathbb{I}[h^{10}_{10}(X) = Y]$ over a large test set.  The correlations ranged from 0.4 to 0.56, with an average value of 0.488.  As a comparison, we also considered a similar calculation for the decomposition produced from a random forest with 500 trees, and from bagging 1000 depth 8 trees.  As with AdaBoost, we can also decompose these classifiers into a sum of interpolating classifiers: 25 sub-ensembles of 20 trees in the case of the random forest, and 10 sub-ensembles of trees in the case of bagged trees.  The interpolating classifiers produced by the random forest had error correlations ranging from 0.75 to 0.87, with an average of 0.81, while the correlations for the bagged trees ranged from 0.9 to 0.94, with an average value of 0.92.  In other words, the sub-ensembles produced by random forests and AdaBoost are less correlated than the bagged trees, with AdaBoost being markedly less so.  When voting across constituents in its decomposition, it is clear that such a decorrelating effect would help to increase the effectiveness of the voting mechanism to cancel out error points.  While we observed this phenomenon in a few data sets, we cannot state under what conditions it happens more generally.  However, it is know that boosting generally outperforms bagging, and the trees in bagging will be much more correlated.

\subsection{Comparison to Boosted Stumps}\label{subsec:stumps}
Throughout this paper we have considered AdaBoost with large trees
of up to $2^8$ terminal nodes.  We have shown that AdaBoost with
such large trees is a classifier which interpolates
the data in such a way that it performs well out of sample for
problems in which the Bayes' error rate is substantially larger
than zero.  In this section we will consider the performance of
AdaBoost with stumps as the base
learner.  The statistical theory predicts that AdaBoost with stumps will overfit less
than AdaBoost with larger trees, as expressed, for instance, in \citet{Jiang2002}. It
is also thought that stumps should be preferable when the Bayes'
decision rule is additive in the original predictors.  For
instance  the seminal book by  \citet[chapter 10]{Hastie2009} 
advocates using trees of a depth one greater than the dominant level of interaction, which is generally quite low.

AdaBoost with stumps does not  self-smooth  nearly as well as
with larger trees, likely because the
the classifiers in the decomposition are more highly correlated, and the ``rough" fits from stumps fail to interpolate the training data locally enough;  the fit is not spiked-smooth around the training set error points. Consequently,  AdaBoost with stumps as base learners  is outperformed by AdaBoost with large trees as base learners, when the Bayes error rate is high. This is the case even when the Bayes rule is
additive.  This result is matched by random forests which works best with large trees. Its randomly chosen predictors at each splitting opportunity lowers the correlation among the trees and the resulting fit is more spiked-smooth.

\begin{figure}[htp]
\centering
    \begin{subfigure}[b]{0.48\textwidth}
        \includegraphics[width=\textwidth]{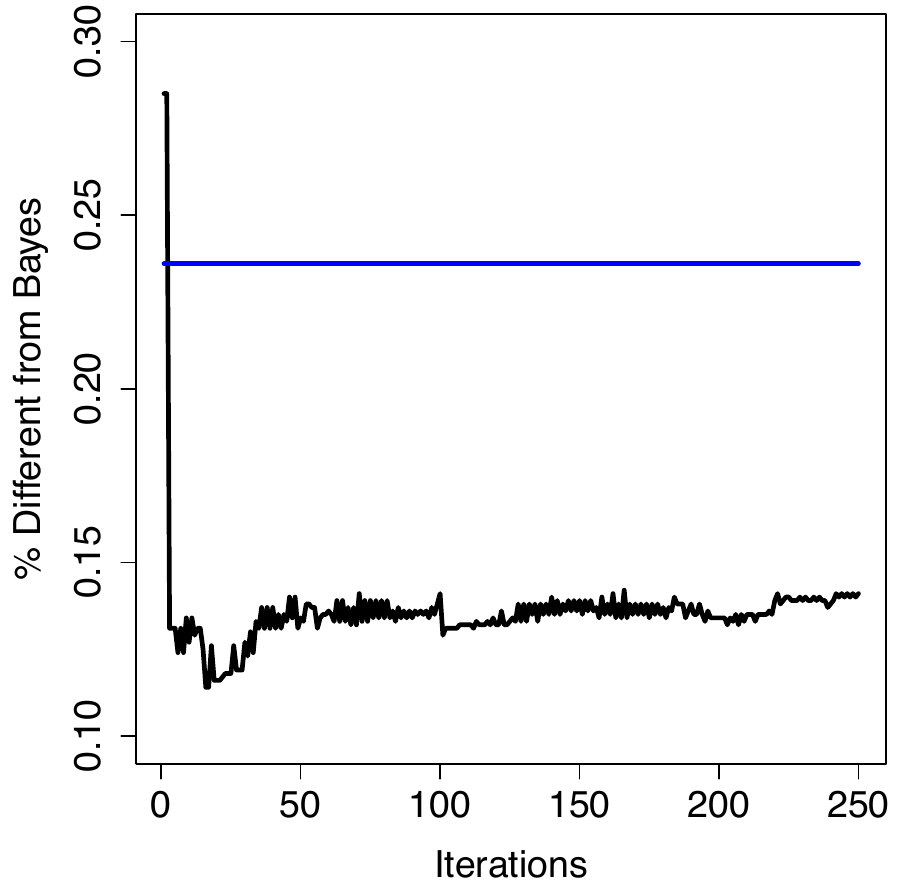}
        \caption{Stump}
	\label{subfig:stump}
     \end{subfigure}
     \quad
     \begin{subfigure}[b]{0.48\textwidth}
        \includegraphics[width=\textwidth]{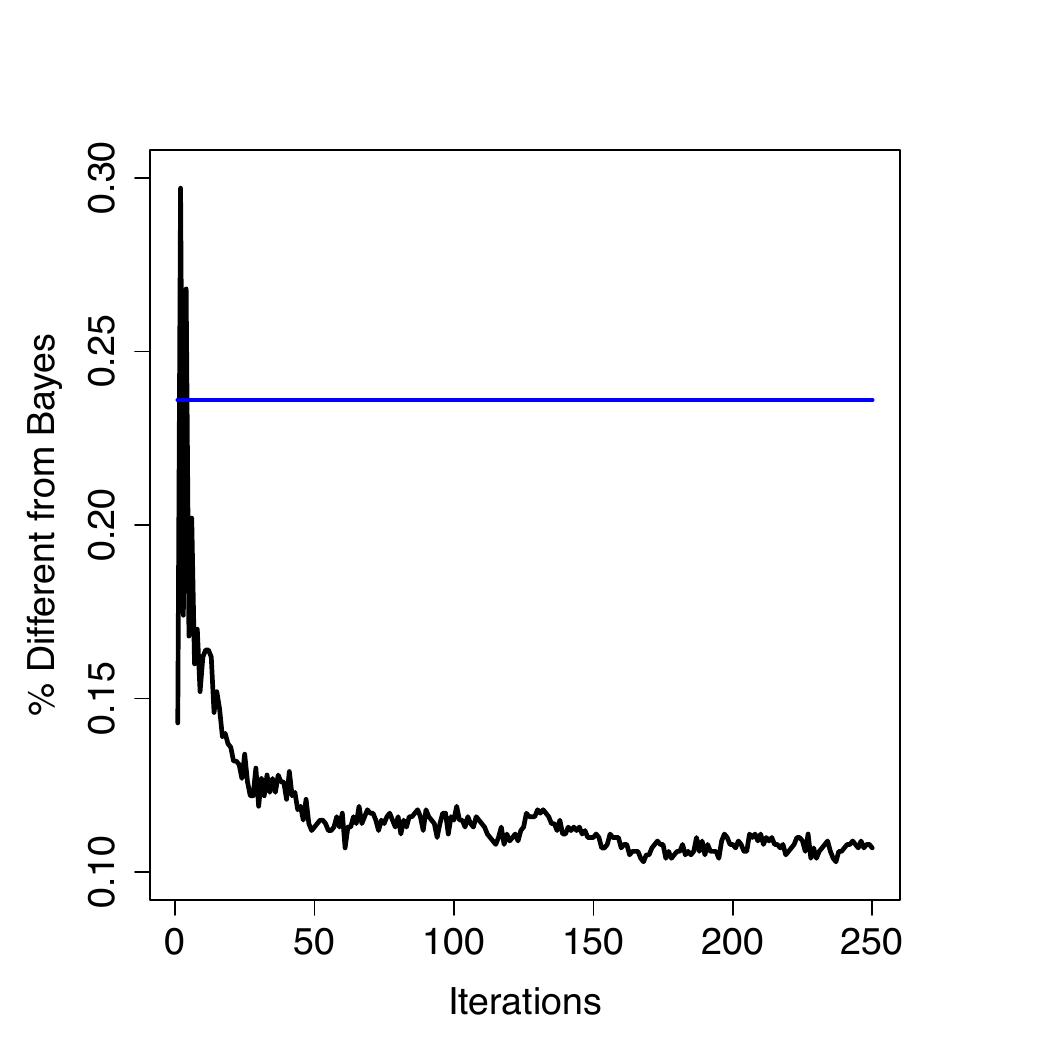}
        \caption{$2^8$ Node Trees}
	\label{subfig:depth8}
     \end{subfigure}
      \caption{Comparison of AdaBoost with stumps (a) and $2^8$ node trees (b) for the five dimensional simulation.  The blue line corresponds to One-NN.} 
\label{fig:comparea}
\end{figure}

To illustrate this with an example,  we return to the five dimensional  simulation from
Section \ref{subsec:five_dim} which has an additive Bayes decision rule.  Figure~
\ref{fig:comparea} displays the percentage of points that are
classified differently from the Bayes rule in the hold out sample
of 1000, as a function of the number of iterations. It can be seen
that the stumps (left panel) do not perform as well as the $2^8$
node trees (right panel).  After 250 iterations, AdaBoost with
stumps yields $141/1000$ points in the hold out sample classified
differently from the Bayes rule, compared to $116/1000$ for this
same data set using instead AdaBoost with $2^8$ node trees.  In
fact, the stumps seems to suffer from overfitting when run beyond
only 25 iterations, while the $2^8$ trees do not have a problem
with overfitting as the number of iterations are increased.


These numerical values are based only on a single run of this
simulation, but the qualitative finding is reproducible over
repeated runs. The result serves to illustrate that the good
out-of-sample performance of AdaBoost using large trees resulting
from the local interpolation of noise points is not shared by
AdaBoost using stumps.  The idea that stumps will perform better
in noisy environments because they overfit less is not supported
by this simulation.  While the stumps do overfit less on the
training data, as evidenced by the fact that they did not give
zero training error, they actually overfit worse than the larger
trees out of sample.  Again, we attribute this to boosting with
stumps lacking the self-smoothing property and not being flexible
enough to interpolate the noise locally.

\section{Real Data Example} \label{sec:realData}

In previous sections, we explored the mechanism of spiked smoothing in simulation experiments.  Here, it was easy for us to identify noise points.  Recall that with fully specified probability model, noise points are simply sample points whose sign differs from the Bayes' rule.  While simulated examples are certainly good for illustration, one may wonder whether these settings are overly simplistic.  Data in the real world is typically generated by more complicated---and unknown---probability models with heteroskedastic noise components.  Since boosted trees have been empirically successful in such settings, it pays to move our discussion to the context of a data set arising in the real world.

\subsection{Phoneme Data}
\label{subsec:phoneme}

In this section, we will discuss spiked smoothing in the context of a data set arising in a speech recognition task designed to discriminate ``nasal" vowels from ``oral" vowels in spoken language \footnote{The original data can be found at the following address: https://www.elen.ucl.ac.be/neural-nets/Research/Projects/ELENA/databases/REAL/phoneme/phoneme.txt.}.  For a collection of 5404 examples of spoken vowels, this data set contains the amplitudes of the first 5 harmonic frequencies as collected by a cochlea spectra (hearing aid), along with a label of either nasal or oral (n=5404, p=5).  The goal is then to use these harmonic frequencies to correctly identify the type of vowel.  While there are many potential data sets we could have considered, this one is convenient to analyze since it consists of a relatively small number of real valued covariates.  This makes it easier to talk about ``neighborhoods" of points later in our discussion.

\begin{figure}[ht!]
\centering
\includegraphics[scale=0.45]{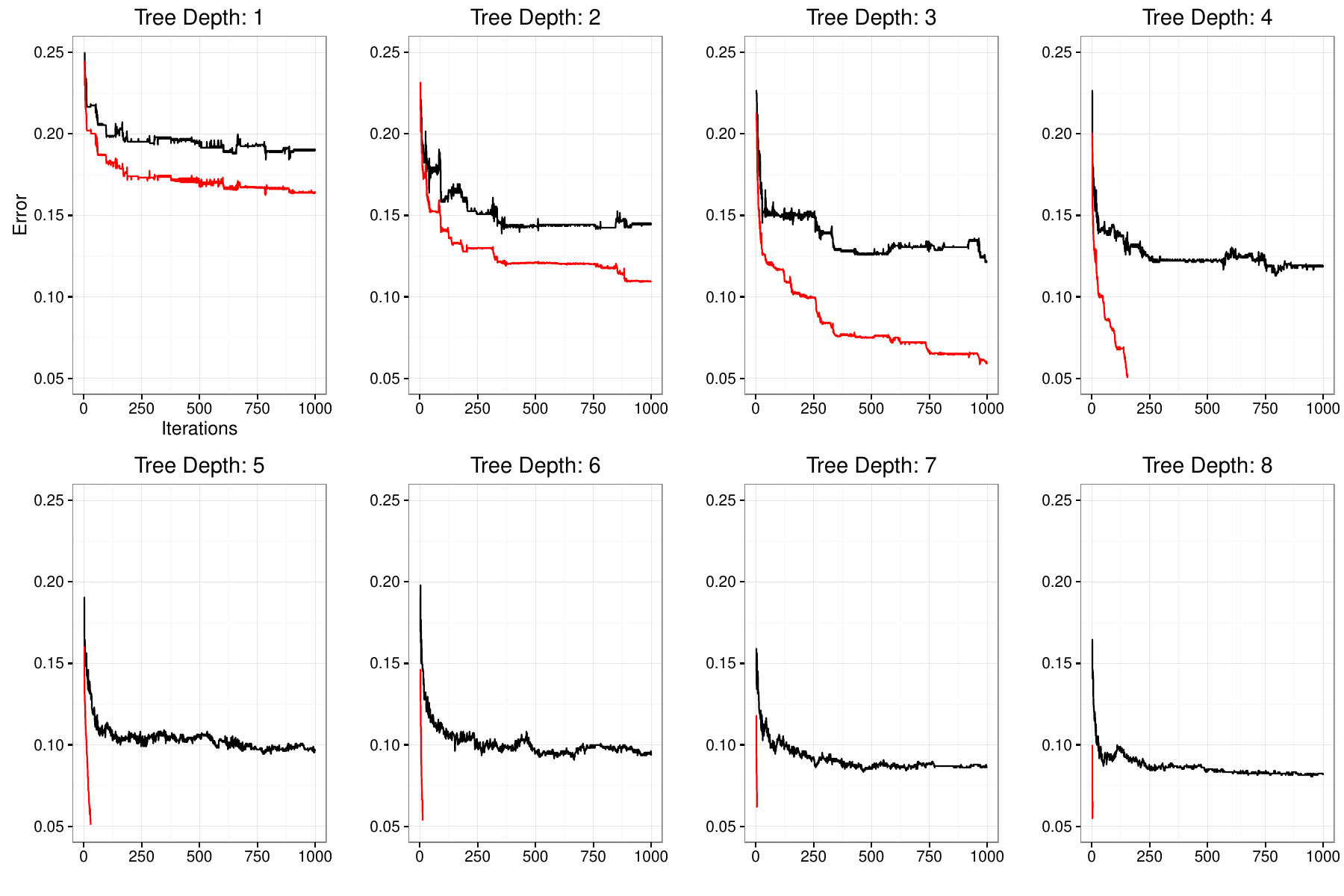}
\caption{Plots of testing and training error as a function of number of boosting iterations for trees of different depths.  The black lines show the test error rates while the red lines show training error rates.  Notice that depth seven and eight boosted trees quickly fit the training data, as depicted by the rapidly decreasing red lines.}
\label{fig:phenomeBoostingDepth}
\end{figure}

We begin by dividing the phoneme data set up into a training set consisting of 70\% of the samples, and a testing set consisting of 30\% of the examples.  Depth eight decision trees boosted for 1000 rounds and a random forest classifier both achieve comparable test error rates of 9.0\% and 9.4\%, respectively \footnote{The error rates reported are the result from repeating the fitting procedure on 100 random train/test set splits and considering the average error on the test set.}.  Figure ~\ref{fig:phenomeBoostingDepth} demonstrates that boosting deep trees is preferable to shallow trees in this data set.  Each frame in the figure shows the testing error in black and the training error in red as a function of the number of boosting iterations for different depth trees.  One can readily observe that testing error steadily decreases with the depth of tree used in AdaBoost.  It is interesting to note that boosting trees of all depths are slow to overfit the data, even with boosted stumps after 1000 iterations.  It is also worth noting that boosted depth 8 trees quickly interpolate the training data, as indicated by the sharply decreasing red line.  This raises the often asked question: what is AdaBoost doing after it achieves a perfect fit on the training set?

\begin{figure}[ht!]
\centering
\includegraphics[scale=0.4]{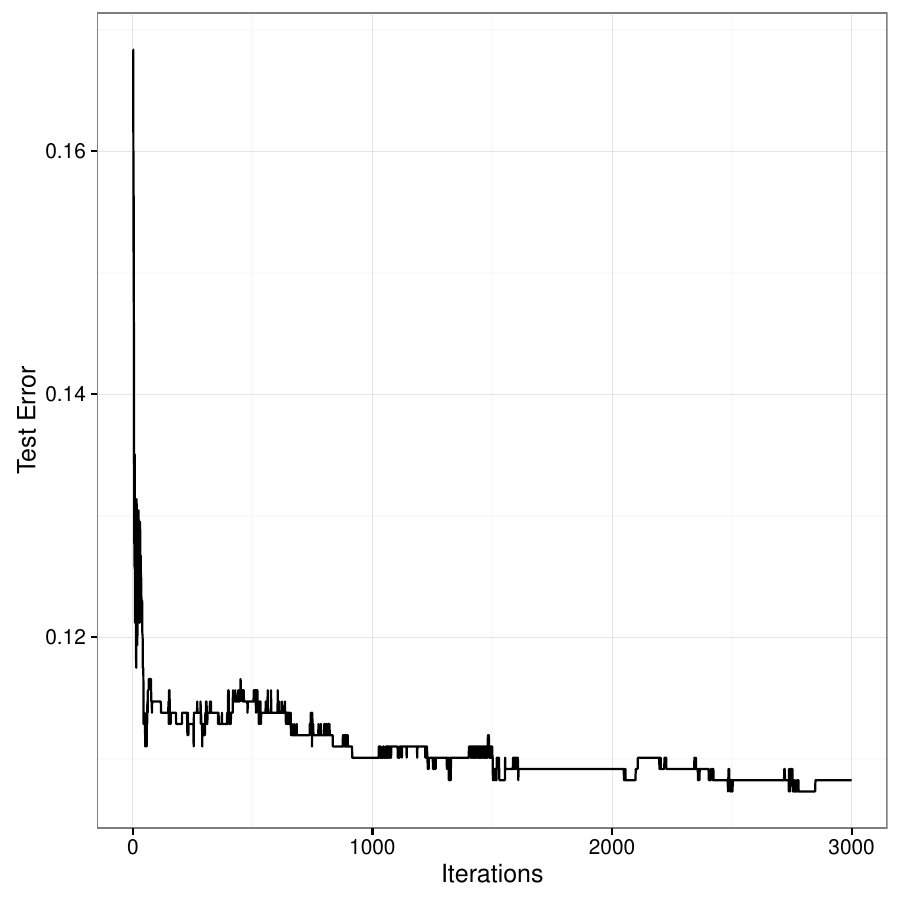}
\caption{Plot of testing error for depth 8 boosted trees fit to a training set which has flipped labels for 100 randomly selected examples.}
\label{fig:phenomeNoisyBoost}
\end{figure}

Any discussion of spiked smoothing on  real data is complicated by the reality that it is impossible to identify noise points without knowing the underlying probability model.  As a substitute for noise points, we flipped the signs of 100 randomly chosen points in our training data set (about 2\% of the data) and analyzed the fits of boosted trees and a random forest classifier around these points.  After refitting the models to the perturbed data, we find the the testing errors are 10.2\% and 10.0\% for boosted depth eight trees and a random forest, respectively.  Figure~\ref{fig:phenomeNoisyBoost} illustrates that even after thousands of rounds of boosting, test error continues to decrease.  The punchline is that both algorithms are able to achieve fits with comparable test set error even after flipping the sign of a large number of points in the training set.  

As a comparison with another interpolating classifier, we also repeated the same experiment with a one nearest-neighbors classifier.  We found the one nearest neighbor achieved a test error rate of 10.5\% when fit on the original training set, and a test error rate of 12.6\% when fit on the noisy training set.  The increase in error rate when noise is added to the training set is larger than that of AdaBoost or a random forest,  which coincides with the results form the simulated example in section \ref{subsec:2d_noise}.  Furthermore, two sample t-tests reveal that the increase in error rate for one nearest-neighbors is significant at the 0.01 level in both cases (p-values less than $1e^{-16}$ in both cases).

\begin{figure}[ht]
\centering
    \begin{subfigure}[b]{0.3\textwidth}
        \includegraphics[width=\textwidth]{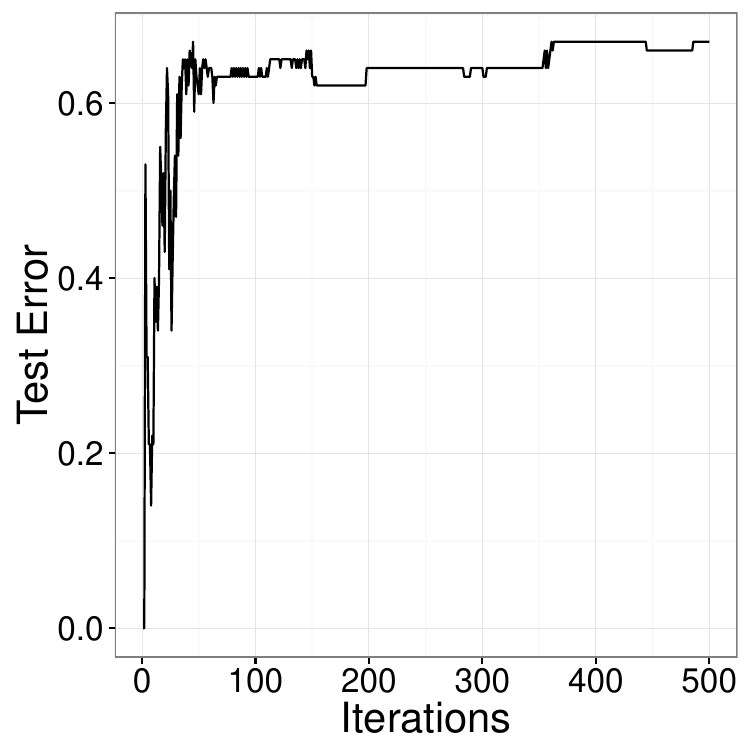}
        \caption{}
     \end{subfigure}
     \quad
     \begin{subfigure}[b]{0.3\textwidth}
        \includegraphics[width=\textwidth]{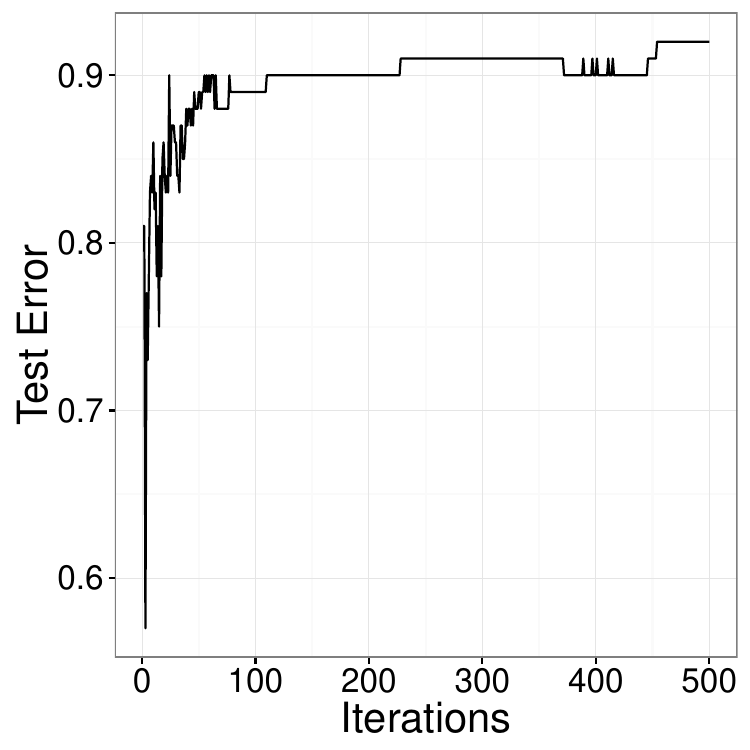}
        \caption{}
     \end{subfigure}

\caption{The fraction of points in a neighborhood that agree with the sign of the original training point before its sign was flipped.  Figures (a) and (b) plot this proportion for two different, representative noise points.}
\label{fig:nhoodNoise}
\end{figure}

We will argue that additional rounds of boosting are helping to ``smooth out" and ``localize" the influence of the 100 noise points that we introduced into our training set.  In section~\ref{subsec:2d_noise}, we illustrated the action of spiked smoothing in diagrams which showed AdaBoost's fit on a two dimensional plane.  In this slightly higher dimensional example, we instead consider small neighborhoods around each of the noise points and track the fraction of points in these neighborhoods that agree with the sign of the ``correct label", that is,  before flipping the sign. \footnote{We choose each neighborhood to be a rectangle centered at the point of interest, with side lengths chosen in such as way that the neighborhood contains only a small number of training points.  We then chose 100 points uniformly at random in this rectangle and computed the AdaBoost classifier at each point: these points are obviously not included in the original data set.}.  Panels (a) and (b) of Figure~\ref{fig:nhoodNoise} plot this fraction as a function of the number of boosting iterations for two representative ``noise points."  In both panels, it is clear that as the number of  iterations increases, the fraction of points in each neighborhood that agrees with the original sign of the training point increases.  Recall that in this case AdaBoost still fits its training data perfectly: the in-sample AdaBoost fit agrees with the sign of the flipped training point in both figures.  Despite this, the algorithm still fits a majority of points in a neighborhood of each of these noise points in the correct way.  The classifier is producing a spiked smooth fit, that is, it fits the data in such a way to localize the influence of noise points.  One can interpret the increasing homogeneity in each neighborhood as the result of averaging.  The first few iterations of deeply boosted trees produce a fit that interpolates the training data, but this fit is quite complicated in the sense that it assigns large numbers of points in the neighborhood to both `+1` and `-1`.  As the number of iterations increases, this fraction increases so that the fit becomes more smooth in the classical sense.

\subsection{Additional Data Sets}
\label{subsec:additional}

We repeated a version of the analysis conducted in Section~\ref{subsec:phoneme} for five additional data sets from the UCI repository: Haberman, Wisconsin breast cancer, voting, Pima, and German credit.  A few notes of the analysis are worth mentioning: we considered adding 5\% label noise, missing values were mean imputed before model fitting, and the experiment was conducted on 50 random training-testing splits of each data set.  Table~\ref{table1} reports the mean increase in testing error after adding 5\% label noise over the 50 random training-testing splits.  For example, on the original haberman data set, AdaBoost achieved an average error rate of 34.225\%, and on the noisy version of the data set achieved an average error rate of 34.354\%.  The mean difference of 0.13\% is reported in the table.  It is also apparent in each setting that AdaBoost and random forest are relatively immune to the addition of label noise.  The stars in the Table~\ref{table1} report the significance level when comparing the increase in error rate for AdaBoost and a random forest with that of one-nearest neighbors using a two-sample t-test.  With the exception of the German credit data, the increase in error rate for one nearest neighbors was larger than that of one nearest neighbors with statistical significance.

\begin{table}[ht]
\centering
\begin{tabular}{l|lll}
  \hline
 Data set & AdaBoost & Random Forest & 1-NN \\ 
  \hline
Haberman & $0.13^{\textbf{**}}$ & $0.52^{\textbf{*}}$ & $\textbf{1.55}^{\textbf{}}$ \\ 
  breast\_cancer & $0.20^{\textbf{***}}$ & $0.39^{\textbf{***}}$ & $\textbf{2.29}^{\textbf{}}$ \\ 
  voting & $1.63^{\textbf{**}}$ & $0.30^{\textbf{***}}$ & $\textbf{2.71}^{\textbf{}}$ \\ 
  Pima & $0.56^{\textbf{***}}$ & $0.45^{\textbf{***}}$ & $\textbf{1.75}^{\textbf{}}$ \\ 
  German & $0.29^{\textbf{}}$ & $0.68^{\textbf{}}$ & $\textbf{0.68}$ \\ 
   \hline
\end{tabular}
\caption{The increase in average testing error after changing the sign of 5\% of the training data.  The stars in the table report the significance level when comparing the increase in error rate for AdaBoost and a random forest with that of one-nearest neighbors using a two-sample t-test.  One star denotes significance at the 10\% level, two stars denotes significance at the 5\% level, and three stars denotes significance at the 1\% level.}
\label{table1}
\end{table}

\section{Concluding Remarks}\label{sec:conclusion}


AdaBoost is an undeniably successful algorithm and random forests is at least as good, if not better. But AdaBoost is as puzzling as it is successful; it broke the basic rules of statistics by iteratively fitting  even noisy data sets until every training set data point was fit without error. Even more puzzling,  to statisticians at least, it will continue to iterate an already perfectly fit algorithm which lowers generalization error.  The statistical view of boosting understands AdaBoost to be a stage wise optimization of an exponential loss, which suggest (demands!) regularization of tree size and control on the number of iterations.  In contrast, a random forest is not an optimization; it appears to work best with large trees and as many iterations as possible. It is widely believed that AdaBoost is effective because it is an optimization, while random forests works---well because it works. Breiman conjectured that ``it is my belief that in its later stages AdaBoost is emulating a random forest" \citep{Breiman2001}.   This paper sheds some light on this conjecture by providing a novel intuition supported by examples to show how AdaBoost and random forest are successful for the same reason.

A random forests model is a weighted ensemble of interpolating classifiers by construction. Although it is much less evident,  we have shown that AdaBoost is also a weighted ensemble of interpolating classifiers.  Viewed in this way, AdaBoost is actually a ``random"   forest of forests.  The trees in random forests and the forests in the AdaBoost each interpolate the data without error. As the number of iterations increase the averaging of decision surface because smooths but nevertheless still interpolates. This is accomplished by whittling down the decision boundary around error points.  We  hope to have cast doubt on the commonly held belief that the later iterations of AdaBoost only served to overfit the data.  Instead, we argue that these later iterations lead to an ``averaging effect", which causes AdaBoost to behave similarly to random forests.

A central part of our discussion also focused on the merits of interpolation of the training data, when coupled with averaging.  Again, we hope to dispel the commonly held belief that interpolation always leads to overfitting.  We have argued instead that fitting the training data in extremely local neighborhoods actually serves to prevent overfitting in the presence of averaging.  The local fits serve to prevent noise points from having undue influence over the fit in other areas.  Random forests and AdaBoost both achieve this desirable level of local interpolation by fitting deep trees.  It is our hope that our emphasis on the ``self-averaging" and interpolating aspects of AdaBoost will lead to a broader discussion of this classifier's success that extends beyond the more traditional emphasis on margins and exponential loss minimization.

\newpage

\bibliographystyle{plainnat}\bibliography{boosting_refs}

\end{document}